\documentclass[journal]{IEEEtran}
\usepackage{amsmath}
\usepackage{graphicx}
\usepackage{multirow}
\usepackage{cite}
\usepackage{setspace}
\usepackage{array}
\usepackage{graphics}
\usepackage{balance}
\usepackage{hhline}
\usepackage{url}
\usepackage{hyperref}
\hypersetup{colorlinks,urlcolor=blue,linkcolor=black,citecolor=black}
\usepackage[font=small]{caption}
\usepackage{subcaption}
\usepackage[export]{adjustbox}
\usepackage{xcolor}
\usepackage{nicefrac}
\usepackage{gensymb}
\usepackage{color}

\makeatletter

\newcommand{\Rmnum}[1]{\expandafter\@slowromancap\romannumeral #1@}

\newlength\replength
\newcommand\repfrac{.33}

\newcommand\rulewidth{.6pt}
\newcommand\tdashfill[1][\repfrac]{\cleaders\hbox to \replength{%
  \smash{\rule[\arraystretch\ht\strutbox]{\repfrac\replength}{\rulewidth}}}\hfill}

\newcommand\tdotfill[1][\repfrac]{\cleaders\hbox to \replength{%
  \smash{\raisebox{\arraystretch\dimexpr\ht\strutbox-.1ex\relax}{.}}}\hfill}
\newcommand\tabdotline{%
  \makebox[0pt][r]{\makebox[\tabcolsep]{\tdotfill\hfil}}\tdotfill\hfil%
  \makebox[0pt][l]{\makebox[\tabcolsep]{\tdotfill\hfil}}%
  \\[-\arraystretch\dimexpr\ht\strutbox+\dp\strutbox\relax]%
}

\makeatother

\ifCLASSINFOpdf
\else
\fi
\begin{document}

\title{Cloud and Cloud Shadow Segmentation for Remote Sensing Imagery via Filtered Jaccard Loss Function and Parametric Augmentation}

\author{Sorour~Mohajerani,~\IEEEmembership{Student Member,~IEEE,}
Parvaneh~Saeedi,~\IEEEmembership{Member,~IEEE}\vspace{-5mm}
\thanks{S. Mohajerani and P. Saeedi are with the School of Engineering Science, Simon Fraser University, BC, Canada (e-mail: smohajer@sfu.ca, psaeedi@sfu.ca).}
}

\maketitle


\begin{abstract}
Cloud and cloud shadow segmentation are fundamental processes in optical remote sensing image analysis. Current methods for cloud/shadow identification in geospatial imagery are not as accurate as they should, especially in the presence of snow and haze. This paper presents a deep learning-based framework for the detection of cloud/shadow in Landsat 8 images. Our method benefits from a convolutional neural network, Cloud-Net+ (a modification of our previously proposed Cloud-Net~\cite{myigarss}) that is trained with a novel loss function (Filtered Jaccard Loss). The proposed loss function is more sensitive to the absence of foreground objects in an image and penalizes/rewards the predicted mask more accurately than other common loss functions. In addition, a sunlight direction-aware data augmentation technique is developed for the task of cloud shadow detection to extend the generalization ability of the proposed model by expanding existing training sets. The combination of Cloud-Net+, Filtered Jaccard Loss function, and the proposed augmentation algorithm delivers superior results on four public cloud/shadow detection datasets. Our experiments on Pascal VOC dataset exemplifies the applicability and quality of our proposed network and loss function in other computer vision applications.

\end{abstract}


\IEEEpeerreviewmaketitle

\vspace{-4mm}\section{Introduction}
\IEEEPARstart{C}{loud} and cloud shadow detection, along with cloud coverage estimation, are among the most critical processes in the analysis of remote sensing imagery. On the one hand, transferring remotely sensed data from air/space-borne sensors to ground stations is an expensive process from time, bandwidth, storage, and computational points of view. On the other hand, no useful information about the Earth's surface can be extracted from optical images that are heavily covered by clouds and their shadows. Since, on average, $67\%$ of the Earth surface is covered by clouds at any given time \cite{cloud_free_sky}, it seems that a considerable amount of resources can be saved by transferring only images  with no/minimum amount of cloud and shadow coverage.

Cloud coverage by itself could provide useful information about climate and atmospheric parameters \cite{cloud_morph}, as well as natural disasters such as hurricanes and volcanic eruptions \cite{volcano, hurricane}. As a result, the identification of clouds and cloud shadows in images is an essential pre-processing task for many applications. Cloud and cloud shadow detection is even more challenging when only a limited number of spectral bands are available. Many air/spaceborne systems such as ZY-3, HJ-1, and  GF-2 are equipped only with visible and near-infrared bands \cite{all_chinese_sats}. Therefore, algorithms that can identify clouds and their shadows from those few spectral bands become more essential.

In recent years, many cloud/shadow detection algorithms have been developed. These methods can be divided into three main categories: threshold-based \cite{thresholdbased8, thresholdbased9, thresholdbased11,thresholdbased13}, handcrafted \cite{hot, handcraft11, handcraft12, handcraft13}, and deep learning-based methods \cite{remotesecnn4, remotesecnn5, remotesecnn6,remotesecnn7, multilevel, remotesecnn12}. 

Function of mask (FMask) \cite{fmask1, fmask2, fmask4} and automated cloud-cover assessment (ACCA) \cite{acca} algorithms are among the most well-known threshold-based algorithms for cloud identification. They construct a decision tree to label each pixel as cloud or other non-cloud classes. In each branch of the tree, a decision is made based on a thresholding function that utilizes one or more spectral bands of the data. 

A group of handcrafted methods isolated haze and thick clouds from other pixels using the relationship between spectral responses of the red and blue bands. Some examples of these algorithms include~\cite{hot2, hot3}, which are derivates of the Haze Optimized Transformation (HOT)~\cite{hot}. Xu et al. \cite{temporal4}, proposed a handcrafted approach, which involved a Bayesian probabilistic model incorporated with multiple spectral, temporal, and spatial features to separate cloud from non-cloud regions.

With recent advances in deep-learning algorithms for image segmentation, several methods have been developed to detect cloud/shadow using deep-learning. Xie et al. \cite{multilevel} trained a convolutional neural network (CNN) from multiple small patches. Their network classified each patch into one of the three classes of thin cloud, thick cloud, and clear. A major issue in cloud/shadow detection approaches using deep-learning is the lack of accurately annotated training images since creating ground truths (GTs) for remote sensing imagery is time-consuming and tedious~\cite{remotesecnn11}. In addition, default cloud masks provided in remote sensing products are mostly obtained through automatic/semi-automatic thresholding-based approaches, which often make them less accurate. Authors in \cite{mymmsp} have removed wrongly labeled icy/snowy regions in default cloud masks in their dataset. They showed that the performance of a well-known semantic segmentation model (U-Net~\cite{unet}) is considerably improved by training on those corrected GTs. Although existing methods deliver promising results, many of them do not provide robust and accurate cloud/shadow masks in scenes where bright/cold non-cloud regions exist alongside clouds \cite{remotesecnn4, fmask2}.

Here, inspired by advances made in deep learning techniques, we propose a new algorithm to identify cloud/shadow regions in Landsat 8 images. Our proposed algorithm consists of a fully convolutional neural network (FCN), which detects cloud and cloud shadow pixels in an end-to-end manner. Our network, Cloud-Net+, is a modified version of our previously open-sourced model, Cloud-Net~\cite{myigarss}. In the process of optimizing Cloud-Net+ model, a novel loss function named Filtered Jaccard Loss (FJL) is developed and used to calculate the error. As a result, Cloud-Net+ is trained more accurately. FJL makes a considerable difference in the performance of systems, especially for images with no cloud/shadow regions. 

Many of the threshold-based and handcrafted algorithms for cloud shadow detection utilize geometrical relationships between the illumination source, clouds, and shadows~\cite{fmask1,shadow_1,handcraft10}. However, such relationships are not taken advantage of in deep learning-based approaches, where shadows are considered independent foreground objects. We incorporate this information through a meaningful parametric data augmentation approach to have a greater variety of shadows in each scene. As a result, newly generated images resemble original images as if they were captured at different times of the day with different sunlight directions. Our experiments show that such systematic augmentation works very well for shadow detection and simultaneous multiclass segmentation of clouds, shadows, and clear areas.

Unlike FMask and ACCA, the proposed approach is not blind to the existing global and local cloud/shadow features of an image. Also, since only four spectral bands---red, green, blue, and near-infrared (RGBNir)---are required for the system's training and prediction, the proposed method can be easily utilized for images obtained by most of the existing satellites as well as airborne systems. Unlike multitemporal-based methods such as \cite{gonzalo_multitemporal}, the proposed method does not require prior knowledge of the scene, e.g., cloud-free images. 
The only data required for the parametric augmentation are solar angles, which exist in the metadata accompanying each image. Moreover, such information is used for training purposes only and not for prediction. In addition, to being simple and straightforward, the proposed network can be used in other image segmentation applications. 

In summary, the contributions of this work are as follows:
\begin{itemize}

\item Proposing a novel loss function (FJL), which not only penalizes a model for poor prediction of clouds and shadows but also \textit{fairly} rewards the correct prediction of clear regions. Our experiments show that FJL outperforms other commonly used loss functions. In addition, when Cloud-Net+ is trained with it, its results outperform state-of-the-art cloud/shadow detection methods over four public datasets of 38-Cloud, 95-Cloud, Biome 8, and SPARCS.
\item Proposing a Sunlight Direction-Aware Augmentation (SDAA) technique to boost the shadow detection performance. Unlike commonly used transformation-based augmentation techniques such as rotation and flipping, which are blind to the scene's geometry, SDAA generates synthetic shadows with various lengths,  directions, and levels of shade.
\item Extending our public dataset, 38-Cloud \cite{myigarss}, for cloud detection in Landsat 8 imagery to a new dataset (95-Cloud). This new dataset---which is made public---includes $57$ more Landsat 8 scenes (than 38-Cloud) along with their manually annotated GTs. It will help researchers improve their cloud detection algorithms using more training and evaluation data.

\end{itemize}

The remainder of this paper is organized as follows: in Section \Rmnum{2}, a summary of related works in the cloud detection field is reviewed. In Section \Rmnum{3}, our proposed method is explained. In Section \Rmnum{4}, experimental results are discussed. Finally, Section \Rmnum{5} summarizes our work.

\vspace{-2mm}\section{Related Works}
Cloud and shadow detection in remotely sensed images have been an active area of research for many years. One of the first attempts to distinguish between clouds and cloud-free areas was through a probabilistic classifier model~\cite{first_cloud_classification}. However, it was limited to scenes with clouds over open sea surfaces. One of the first  successful and more general automatic cloud detection methods was Fmask (first version) \cite{fmask1}. Fmask used seven bands of Landsat images to classify each pixel of an image into one of the five classes of land, water, cloud, shadow, and snow. Another version of it (Fmask V3) \cite{fmask2} utilized cirrus band to distinguish cirrus clouds along with low altitude clouds. In the last version of it (Fmask V4) \cite{fmask4}, auxiliary data such as digital elevation maps (DEM), DEM derivatives, and global surface water occurrences (GSWO) were used in addition to the other usual bands for better performance on water, high altitude regions, and Sentinel-2 images. 

Several multitemporal methods were developed for cloud/shadow detection~\cite{temporal3, temporal4, temporal6, temporal7}. Mateo-Garc\'{i}a et al. \cite{gonzalo_multitemporal} used cloud-free images of Landsat 8 scenes to identify potential clouds. A clustering and some threshold-based post-processing steps then helped to generate final cloud masks. Zi et al. \cite{pcanet} combined a threshold-based method with a classical machine learning approach to segment superpixels and classify them into one of three classes of cloud, potential cloud, and non-cloud.

Recently, several FCNs have been developed for cloud and cloud shadow detection tasks~\cite{remotesecnn9, remotesecnn1, remotesecnn2, remotesecnn3, remotesecnn10, remotesecnn11}. Yang et al. \cite{cdnet} proposed an FCN, which detects clouds in ZY-3 thumbnail satellite images. A built-in boundary refinement approach was incorporated into their network (CDnetV1) to avoid further post-processing. The second version of that network, CDnetV2~\cite{cdnet2}, was equipped with two new feature fusion and information guidance modules to extract cloud features more accurately. Chai et al. ~\cite{remotesecnn8} proposed Segnet Adaption (a modified version of a well-known FCN for semantic segmentation) to match with remotely sensed images. Segnet Adaption has shown promising results on Biome 8 and Landsat 7 cloud/shadow detection datasets. Recently, Jeppesen et al. \cite{rsnet} introduced RS-Net to identify clouds in Landsat 8 images. RS-Net was an FCN inspired by U-Net and was trained with both automatically (via Fmask) and manually generated GT images of two public datasets. The authors showed that results obtained by a model trained with Fmask's outputs outperform the Fmask directly obtained results. 

The authors of SPARCS CNN~\cite{sparcs2} also proposed a model to distinguish cloud, shadow, water, snow, and clear regions in Landsat 8 images. They used a pretrained VGG16 as the backbone of their fully convolutional network. They succeeded in reaching human interpreter accuracy with their model. RefUnet v1~\cite{refunet1} was another method which focuses on retrieving fuzzy boundaries of clouds and shadows. The authors used a UNet for extracting course/rough cloud and cloud shadow in small patches of images. Then, boundaries of clouds/shadows in complete Landsat 8 masks were refined/sharpened using a dense conditional random field (CRF). In this method, the CRF refinement was applied as a post-processing step. In the second version of RefUNet (RefUNet v2)~\cite{refunet2} a simultaneous joint pipeline for detecting and refining edges was  utilized. Therefore, the UNet network, which identified course masks, was concatenated with the proposed Guided Gaussian filter-based CRF to refine boundaries in an end-to-end manner. In addition, refinements were done on small patch masks rather than large masks of the entire Landsat 8 scenes. In another work, the authors of Cloud-AttU~\cite{cloud_attu} employed a UNet model that is enriched with a specific attention module in its skip connections. Multiple attention modules enabled the model to learn proper features by paying attention to the most relevant locations in input training images or feature maps. CloudFCN~\cite{cloudfcn} was based on a UNet model with the addition of inception modules between convolution blocks in the contracting and expanding arms. In addition to pixel-level segmentation, the authors incorporated a feature into their algorithm for estimating cloud coverage in an image using a regression-friendly loss function. They demonstrated that CloudFCN could achieve high accuracy in the task of cloud screening under various conditions, such as the presence of white noise and using different quantization methods.

Image augmentation has been proven to be effective in increasing the accuracy of deep learning models. There are only a few offline augmentation algorithms reported for remote sensing imagery. Ma et al.~\cite{aug_gan1} proposed a generative adversarial network (GAN) to synthesize images for scene labeling. Howe et al.~\cite{aug_gan2} introduced another GAN-based approach for Earth's surface object detection in airborne images. Zheng et al.~\cite{aug_gan3} proposed a method for generating synthetic vehicles in aerial images. For the specific task of cloud/shadow segmentation, only basic geometric and color space transformations in the training phase are used for data augmentation~\cite{shadow_2, remotesecnn5}. Authors of~\cite{myicip2019}, however, developed a GAN (CloudMaskGAN) to convert snowy scenes to non-snowy and vice versa for augmentation of Landsat 8 images. To the best of our knowledge,  CloudMaskGAN is the only reported non-transformation-based augmentation approach for cloud detection purposes. 

\section{Proposed Method}
In this section, the proposed methodology for cloud/shadow detection is described. We focus on using Landsat 8 images. Landsat 8 is equipped with two optical sensors, which together collect eight spectral and two thermal bands. Only four spectral bands---Bands $2$ to $5$ (RGBNir)---are used (all with $30m$ spatial resolution). To keep the proposed method more general, no thermal band is utilized.

\vspace{-3mm}\subsection{Architecture of the Segmentation Model}
Similar to other FCNs, the proposed network is made of two main arms: a contracting arm, and an expanding arm. In the training phase, the contracting arm extracts important high-level cloud/shadow attributes. It also downsamples the input while increasing its depth. On the other hand, the expanding arm utilizes those extracted features to build an image mask---with the same size as the input. The network's input is a multi-spectral image, and its output is a grayscale cloud/shadow probability map for the input. In the case of multiclass segmentation, the output has multiple channels, one corresponding to each class.

\begin{figure}[h]
\begin{minipage}{0.5\textwidth}
\centering
\includegraphics[width = 70mm]{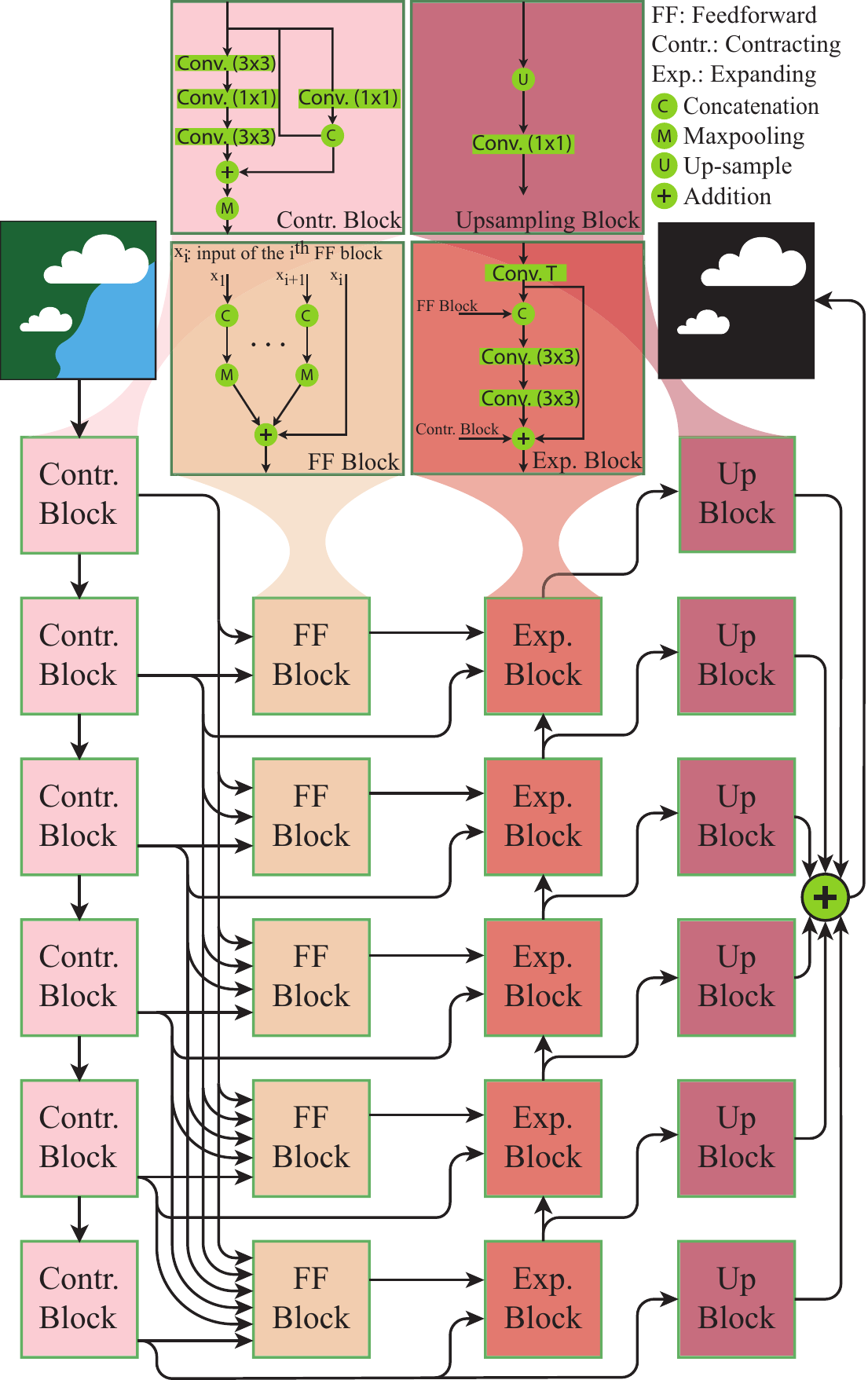}
\end{minipage}
\caption{\small Cloud-Net+'s architecture.   
\label{Fig:arch}}
\vspace{-5.5mm}
\end{figure}

We modified the network's architecture in our previous work \cite{myigarss} to develop a more efficient model that is more sensitive to clouds/shadows. Cloud-Net+, consists of six contracting and five expanding blocks  (see Fig.\ref{Fig:arch}). Successive convolutional layers are the heart of the blocks in both arms. The kernel size and the order of these layers play crucial roles in the quality of activated features, and therefore, they affect the final segmentation outcome directly. On the one hand, it seems that as the number of convolutional layers in each block increases, the distinction of the captured context improves. On the other hand, utilizing more of such layers explodes the complexity of the model. To address such trade-off, we remove the middle $3\times 3$ convolution layer in the last two contracting blocks and the first expanding block of Cloud-Net. Since those layers are dense and contain thousands of parameters, such removal decreases the number of network's parameters significantly. Then, in all contracting arm blocks, a $1\!\!\times\!\!1$ convolution layer is added between each two adjacent $3\!\times\!3$ convolution layers. Since each $1\!\!\times\!\!1$ convolution layer contains a small number of parameters, the total number of parameters of Cloud-Net+ ($32.9$M) is $10\%$ less than that of Cloud-Net ($36.4$M). Utilizing the $1\!\times\!1$ kernel size in convolutional layers was suggested initially in \cite{net_in_net}, and its effectiveness is proven in works such as \cite{wide-res}. Employing such a kernel in expanding blocks does not yield a better recovery of the low-resolution feature maps. Therefore, we do not add it to the expanding blocks. Instead, an aggregation branch (AB) is added to combine all feature maps of the expanding blocks. AB consists of six up-sampling layers (bilinear interpolation) followed by a $1\!\!\times\!\!1$ convolution. It helps our model to retrieve cloud/shadow boundaries in the generated masks. 

\vspace{-3mm}

\vspace{-1mm}\subsection{Loss Function}
Soft Jaccard/Dice loss function has been widely used to optimize many image segmentation models \cite{jacc1,dice1,jacc2}. The formulation of soft Jaccard loss for two classes of "$0$" and "$1$" is as follows:
\vspace{-5mm}\begin{equation}
\small
\begin{split}
J_{L}(t,y) \! = 1 \!-\dfrac{\sum\limits_{i=1}^{N} t_{i} y_i+\epsilon}{\sum\limits_{i=1}^{N} t_{i} + \sum\limits_{i=1}^{N} y_i - \sum\limits_{i=1}^{N} t_{i} y_i+\epsilon},
\\ 
\end{split}
\label{Eq:jloss}
\end{equation}
Here $t$ represents the GT, and $y$ is the output of the network. $N$ denotes the total number of pixels in $t$. $y_i \in [0,1]$ and $t_{i} \in \{0,1\}$ are the $i$th pixel value of $y$ and $t$, respectively, and $\epsilon$ is $10^{-7}$ to avoid division by zero. Soft Jaccard loss function, however, has a defect: over-penalization of images with no class "$1$" in their GTs. 

Let us consider a small $2\times2$ input image with $t_{0} \!=\![0, 0; 0, 0]$ and two possible predictions of $y_1\!=\![0.01, 0.01; 0.01, 0.01]$ and $y_2\!=\![0.99, 0.99; 0.99, 0.99]$. It is clear that $y_1$ would be a better prediction than $y_2$ since it can be interpreted as having no class "$1$" in the input image. However, soft Jaccard losses obtained by $y_1$ and $y_2$ are the same: $J_{L}(t_{0},y_1)  \!= \! J_{L}(t_{0},y_2)\approx 1$. Consequently, the network penalizes $y_1$ as much as $y_2$, even though $y_1$ represents a better prediction. Indeed, the major problem with soft Jaccard loss function is that whenever there is no class "$1$" in the GT, the numerator of Eq. (\ref{Eq:jloss}) equals to $\epsilon$ (which is a small number) and, as a result, the value of the loss approximates $1$.

Observing this behavior, we propose a modified soft Jaccard loss function, FJL, with two  different versions. The main idea behind FJL is to compensate for unfair values of soft Jaccard loss and replace them with proper values whenever there is no class "$1$" in the GT. We can summarize the goal of the FJL as follows:
\vspace{-2mm}\begin{equation}
\small
\mathit{FJL} (t,y) \! =\! 
   \begin{cases} 
    \!  G_{L}(t,y), & \; t_{i} =0, \forall i \in \! \{\! 1,\! 2,\! 3,..., \! N\} \\
    \! J_{L}(t,y), & Otherwise 
   \end{cases}
\vspace{-1.5mm}
\label{Eq:goal_filter}
\end{equation}
where $G_{L}$ represents a compensatory function. The condition in the first line of Eq. (\ref{Eq:goal_filter}) indicates that when all pixels of GT are equal to zero, FJL is similar $G_{L}$. Clearly, this condition can be rephrased as follows:
\vspace{-2mm}
\begin{equation}
\small
\mathit{FJL} (t,y) \! =\! 
   \begin{cases} 
    \! G_{L}(t,y), & S=0 \\
    \! J_{L}(t,y), & S>0 
   \end{cases}
\vspace{-2mm}
\label{Eq:goal_filter_reph}
\end{equation}
where $S\!=\!\sum_{i=1}^{N} \! t_{i}$. This formulation can be rewritten---in a more general form---as the combination of two functions of $J_{L}$ and $G_{L}$, in which each function is multiplied by ideal highpass and lowpass filters, respectively:

\vspace{-5mm}
{\small
\begin{alignat}{1}
& \mathit{FJL} (t,y) \! =   
 k_{G}G_{L}(t,y) LP_{p_{c}}(S)  +   k_{J}  J_{L}(t,y) H\!P_{p'_{c}}(S)
\label{Eq:filtered_j1}
\end{alignat}
}%
Here, $LP_{p_{c}}$ denotes a lowpass filter with the cut-off point of $p_{c}$ and $HP_{p'_{c}}$ denotes a highpass filter with the cut-off point of $p'_{c}$. $k_{G}$  and $k_{J}$ represent coefficients of compensatory and Jaccard losses, respectively.  The magnitude of both filters is limited in the [0,1] range. In Eq. (\ref{Eq:filtered_j1}), the value of $LP_{p_{c}}$ is $0$ when $S \!>\! p_{c} $, so the value of $G_{L}(t,y) LP_{p_{c}}$ becomes zero, and as a result, FJL only has contributions from $J_{L}$. On the other hand, when $S \!<\! p'_{c}$, $H\!P_{p'_{c}}$ becomes $0$, and FJL is only from the $G_{L}$ part. This behavior can be simply described as a toggle switch between $J_{L}$ and $G_{L}$. 

To satisfy Eq. (\ref{Eq:goal_filter_reph}), the cut-off point of the two ideal filters should be equal, so they become complementary. Note that, in signal processing context, the cut-off usually refers to the "frequency" at which the magnitude of a filter changes. However, in this work, cut-off is the "value" of $S$ at which the magnitude of a filter alters. Since this transition should occur when $S=0$, the cut-off points are set to $0$. 

To have lowpass and highpass filters with smooth gradient characteristics, inspiring by \cite{loss_idea}, we have used the sigmoid function as follows:
\vspace{-1mm}\begin{equation}
\small
 \hspace{-3mm} L\!P_{p_{c}}(S) \! =  \!  \frac{1}{ 1 \! + \! exp \big( \! m ( S\! - \!p_{c} ) \!  \big)} , H\!P_{p'_{c}}(S) \! = \! \frac{1}{1 \! + \! exp \big( \! m ( \! - S\! + \!p'_{c} ) \! \big)} \\
\label{Eq:filterlh}
\end{equation}
where $m$ denotes the steepness of the sigmoid transition. Choosing sigmoid solves the problem of over-penalizing without adding any non-differentiable elements or piecewise conditions to the loss function. This leads to a smooth switch between soft Jaccard and compensatory loss in FJL. Since $H\!P$ and $LP$ filters are not functions of $y$, the gradient of FJL is still continuous (without any jumps). By substituting  Eq. (\ref{Eq:filterlh}) in Eq. (\ref{Eq:filtered_j1}), FJL is described as:

\vspace{-5mm}
{\small
\begin{alignat}{1}
  \mathit{FJL} (t,y) \! =  
\frac{ k_{G}G_{L}(t,y)}{ 1 \! + \! exp \big( \! m ( S\! - \!p_{c} ) \!  \big)}   + 
 \frac{k_{J}J_{L}(t,y)}{1 \! + \! exp \big( \! m ( \! - S\! + \!p'_{c} ) \! \big)}
\label{Eq:filtered_j2}
\end{alignat}
}%

To keep these filters close to ideal, $m$ is required to be a large number. We have set $m$ to $1000$ in our experiments to have fast transitions from $0$ to $1$ and vice versa. In addition, $ p_{c}\!$ and $\!p'_{c}$ are set to $0.5$ to leave a safety margin and ensure that $LP_{0.5}(S\!=\!0)\!\!=\!1$ and $HP_{0.5} (S\!=\!0)\!\!=\!0$. Also, by setting $k_{G}\!=\!k_{J}\!= \! 1$, the magnitude of the loss remains in $[0,1]$ range. 

We can have different versions of FJL, by selecting different functions as $G_{L}$. Here, we investigate two different candidates, $G_{L1}$ and $G_{L2}$, to form two corresponding FJL versions, $F\!J\!L_{1}$ and $F\!J\!L_{2}$. Our first candidate is the inverted Jaccard function, which is calculated by the complements of GT and prediction arrays. The second one is the normalized  version of a common loss function in segmentation tasks, Cross Entropy (CE). The formulation of these candidates are as follows:
\vspace{-1mm}\begin{alignat}{2}
\small
& \!\!\! G_{L1}(t,y)\!\equiv\!InvJ_{L}(t,y) \! = J_{L}(\overline{t \vphantom y}, \overline y), \\[-4mm] 
& \!\!\! \nonumber G_{L2}(t,y)\!\equiv\! CE_{norm}(t,y) \!\! = \!\! \frac{CE(t,y)}{Max} \!\!= \!\! \frac{ \dfrac{-1}{N} \! \! \sum\limits_{i=1}^{N} \! \big( t_i  log(y_i \! + \! \epsilon)\! \big)}{Max}   ,
\vspace{-2mm}
\label{Eq:invjloss}
\end{alignat}
where $\overline{t}$ and $\overline{y}$ denote the \textit{complements} of $t$ and $y$, respectively. $CE_{norm}$ denotes normalized CE. To make sure that the range of $F\!J\!L_{2}(t,y)$ is bounded to $[0,1]$, CE values are normalized by division by their maximum possible value, $Max= \frac{-N}{N}\log({\epsilon}) = 16.1180$. $Max$ is obtained when all pixels in the predicted array are complements of GT.

\vspace{-2mm}
\subsection{Parametric Augmentation (SDAA)}\label{SDA}
When an image is captured by Landsat 8, the solar azimuth ($\theta_{A}$) and the zenith ($\theta_{Z}$) angles at the time of acquisition are recorded in a metadata file. We use these angles to generate different synthetic images by hypothetically changing the sun's location in the sky and, therefore, creating shadows under different illumination directions. This could be interpreted as capturing the same scene at different times of the day. Each image can be augmented unless it is free of shadow. 

The main steps of the proposed SDAA are as follows: 1) removing original shadows, 2) changing the direction of illumination, 3) generating the locations of synthetic shadows by projecting clouds using the synthetic illumination direction of step 2, 4) recoloring synthetic shadow locations to generate different levels of shades via gamma transformation. Details of these steps are explained next.

First, a Landsat 8 image (from the training set), its shadow GT, and its cloud GT are selected. To generate realistic images, original shadows should be removed from the image, otherwise, the synthetic image will have two cast shadows. For the shadow removal step, a histogram matching between shadow regions and their shadow-free neighborhood is performed. The result of this step is a shadow-free image.

\vspace{0mm}\begin{figure}[h]
\footnotesize
\hspace{-2mm}\begin{minipage}{0.5\textwidth}
\centering
 \includegraphics[width = 90mm]{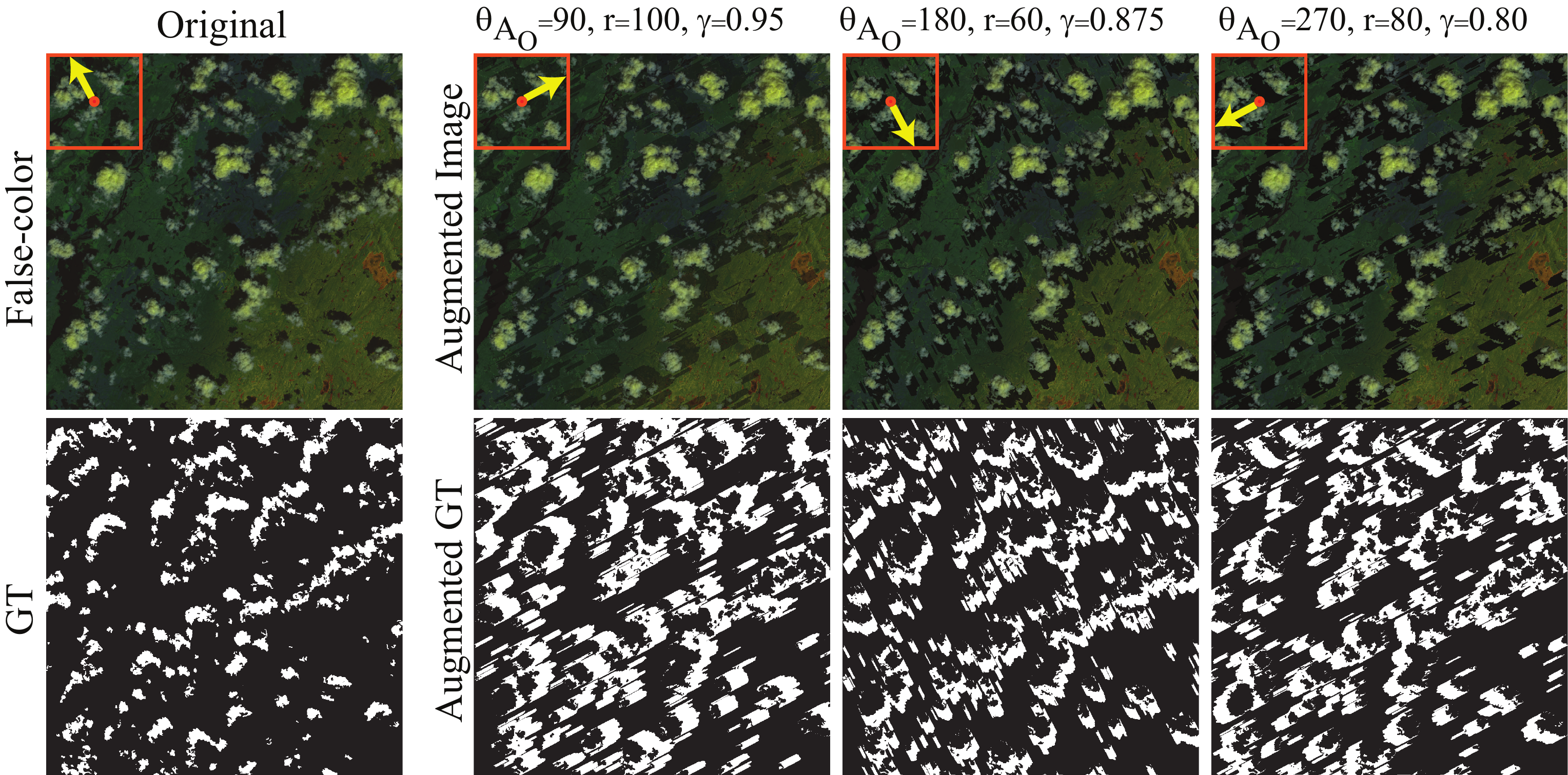}
\end{minipage}
\caption{\footnotesize Examples of SDAA obtained from  one of the SPARCS images.  The yellow arrow shows the sunlight direction in each image.
\label{Fig:sh_aug_exm}}
\vspace{-2mm}
\end{figure}

To create new shadows in the shadow-free image, we generate synthetic solar angles by adding offsets to the original solar angles of $\theta_{A}$ and $\theta_{Z}$. Using these new angles, the direction of synthetic sunlight (in form of a $2D$ vector projected on the scene's image plane) is calculated. By moving cloud pixels on the image plane and in the direction of the new sunlight, synthetic shadows of those clouds are generated. We made the following assumptions to simplify this process: all clouds have the same height, and the scene is located on a flat plane. 
 The following equations are used to obtain the new locations of the cast shadows:

\vspace{-4mm}
\begin{equation}
\small
\begin{split}
y_{sh}\! = y_{cl}\! + \!r \sin(\theta_{Z} + \theta_{Z_O})\cos(\theta_{A} + \theta_{A_O}),\\
x_{sh}\! = x_{cl}\! + \!r \sin(\theta_{Z} + \theta_{Z_O})\sin(\theta_{A} + \theta_{A_O}),
\\ 
\end{split}
\vspace{-2mm}
\label{Eq:fake_angles}
\end{equation}
where $y_{cl}$, $y_{sh}$, $x_{cl}$, and $x_{sh}$ denote the location of a cloud pixel and its corresponding cast shadow along the $y$ and $x$ axes. $\theta_{Z_O}$ and $\theta_{A_O}$ represent  solar angle offsets. $r$ (in pixels) is the shifting factor, which defines the length of a cast shadow. The greater the $r$ is set, the more a piece of cloud is shifted, and therefore, the longer its shadow becomes. Since both $r$ and $\theta_{Z}$ could affect the length of the cast shadow, instead of altering both, we keep $\theta_{Z}$ intact and only alter $r$. Therefore, $\theta_{Z_O}$ is set to $0$.
 
The resultant synthetic shadow mask (SSM) will be used as the shadow GT of the augmented image in the training phase. In the next step, SSM is used to adjust the brightness of the synthetic shadow regions in the augmented image using gamma ($\gamma$) transformation ($i'\!=\!i^{\gamma}$). Pixel values of shadow-free regions in the image---including cloudy and clear areas---are not modified. 

Multiple candidates have been considered for hyperparameters of SDAA: $\theta_{A_O}$, $r$, and $\gamma$. These candidates consist of: $\theta_{A_O} =\{90, 180, 270\}\degree$, $r=\{20, 40, 60, 80, 100\}$, and $\gamma = \{0.8, 0.825, 0.85, 0.875, 0.9, 0.925, 0.95, 0.975\}$. Note that $\theta_{A_O}$ is selected to cover all possible ranges of the solar azimuth angle---$[0, 360]$. Values of $r$ greater than $100$ generate too long and unrealistic shadows. In addition, choosing $\gamma$ values smaller than $0.8$ and greater than $0.975$ leads to too dark and unnatural-looking bright shadows. The entire process is repeated for all spectral bands (R,  G, B, and Nir). Fig. \ref{Fig:sh_aug_exm} displays some of the  augmented images generated by SDAA.

\section{Experimental Settings and Results}

\subsection{Training Details}

The size of the input data for Cloud-Net+ is $192\!\times\!192\!\times\!4$. Four spectral bands of each patch are stacked up in the following order: R, G, B, Nir, to create such input. Before training, input patches are normalized through division by 65535. $20\%$ of training data is used for validation during the training.

Input patches are randomly augmented using simple online geometric translations such as horizontal flipping, rotation, and zooming. Note that to make the proposed loss functions compatible with multiclass segmentation experiments, in each iteration of the training, loss values obtained for each class are averaged (with weights proportional to the inverse number of pixels in each class). The activation function in the last convolution layer of the network is a sigmoid. In the case of multiclass segmentation, softmax function is used in the last layer. Adam method~\cite{ADAM} is utilized as the optimizer. 

The initial weights of the network are obtained by a Xavier uniform random initializer \cite{xavier}. The initial learning rate for the model's training is set to $10^{-4}$. If the validation loss  does not drop for more than 15 successive epochs, the learning rate is reduced by $70\%$. This policy is continued until the learning rate reaches $10^{-8}$. The batch size is set to $12$. The proposed network is developed using Keras deep learning framework with a single GPU.

\subsection{Datasets}
\subsubsection{38-Cloud Dataset}
38-Cloud dataset,  which has been introduced in \cite{myigarss}, consists of $8400$ non-overlapping (NOL) patches of $384\!\!\times\!\!384$ pixels extracted from $18$ Landsat 8 Collection 1 Level-1 scenes as the training set. $9201$ patches of the same spatial size obtained from $20$ Landsat 8 scenes represent the test set. Scenes are mainly from North America and their GTs are manually extracted. This dataset includes only four spectral bands of R, G, B, and Nir. 

All Landsat 8 scenes have black (empty) regions around them. These regions are created when the acquired images are aligned to have the geodetic north at their top. We eliminated training patches with more than $80\%$ empty pixels. The number of non-empty training patches decreases to $5155$, resulting in a significant decrease in training time.

\subsubsection{95-Cloud Dataset}
To improve the generalization ability of deep neural networks trained on 38-Cloud dataset, we have extended it by adding $57$ new Landsat 8 scenes to the training scenes of 38-Cloud dataset. Therefore, in total, the new training set consists of $75$ scenes. 38-Cloud test set has been kept intact in 95-Cloud dataset for evaluation consistency. The GTs for the new scenes are manually extracted. Different images in 95-Cloud are selected to include various land cover types such as soil, vegetation, urban areas, snow, ice, water, haze, and different cloud patterns. The average cloud coverage percentage in  95-Cloud dataset images is kept around $50\%$. Following the same pattern as in 38-Cloud dataset, the total number of patches for training is $34701$ and for the test is $9201$. Removing empty patches from 95-Cloud training set reduces the number of patches to $21502$. We made this dataset and parts of the code publicly available to the community at  \href{https://github.com/SorourMo/95-Cloud-An-Extension-to-38-Cloud-Dataset}{https://github.com/SorourMo/95-Cloud-An-Extension-to-38-Cloud-Dataset}. 

\subsubsection{Biome 8 Dataset} 
Biome 8 dataset~\cite{biome8_1} is a publicly available dataset consisting of 96 Landsat 8 scenes with their manually generated GTs with five classes of cloud, thin cloud, clear, cloud shadow, and empty. We generated a binary cloud GT out of Biome 8 GTs by merging both thin cloud and cloud classes into one cloud class. We marked the rest of the classes as clear. For the shadow segmentation task, all classes, except shadow, are combined under a clear class. Following the same pattern from 38-Cloud and 95-Cloud datasets, the total number of cropped patches extracted from the entire Biome 8 dataset is $\! 44327\!$. Removing empty patches reduces this number to $27358$.

\subsubsection{SPARCS Dataset} {\label{sparcs_dataset}}
SPARCS dataset \cite{sparcs, sparcs2} consists of 80 patches of $1000\! \times \!1000$ extracted from complete Landsat 8 scenes. The GTs of these patches are manually generated. Each pixel is classified into one of the cloud, shadow, snow/ice, water, land, and flooded classes. For the binary segmentation of clouds/shadows, we have combined all non-cloud and non-shadow classes under the clear class. In multiclass segmentation, three classes of cloud, shadow, and clear have been kept. Following the same pattern as previous datasets, the number of extracted NOL patches for this dataset is $720$.

\vspace{-3.5mm}\subsection{Model Evaluation}
To evaluate the performance of the proposed algorithm, various experiments are performed. For 38-Cloud and 95-Cloud datasets, our model is trained  and tested on each of those datasets' explicitly defined training set and testing set, respectively. For SPARCS dataset, as suggested in \cite{rsnet}, we randomly extract five folds of images (each fold consists of 16 complete images or 144 NOL patches). Folds 2 to 5 are considered for training and fold 1 for testing in our experiments. We have also conducted 5-fold cross-validation (5CV) over the extracted folds to compare obtained results with those from RS-Net~\cite{rsnet}. 

For Biome 8 dataset, we follow the instructions described in \cite{rsnet} and extract two folds from Biome 8 dataset. We randomly select two cloudy, two mid-cloud, and two clear scenes for each biome category. Therefore, 48 scenes are extracted for fold 1 and 48 scenes for fold 2. In our experiments, fold 2 is used for training and fold 1 for testing. To compare against RS-Net, 2-fold cross-validation (2CV) is conducted, and the obtained numerical results are averaged.

Evaluation of the SDAA method requires different arrangements. For SPARCS, multiple sets of augmented training images are generated using various combinations of hyperparameters described in section \ref{SDA}. NOL patches of $384\!\times\!384$ pixels are extracted from those images and are added to the original training patches of SPARCS dataset one set at a time. Then, to find the best combination of hyperparameters, shadow detection training and testing are performed for each of those expanded training sets. The numerical results of these experiments are calculated and compared against those obtained from training with original patches only. Those combinations of hyperparameters that resulted in a higher Jaccard index than the training with original patches are identified (11 combinations). SDAA patches are generated by those combinations and added all together to the original training patches of a dataset for a final evaluation.

Considering SPARCS's folds 2 to 5 as the training set, we use the best hyperparameters confirmed in the previous steps to generate SDAA patches, leading to $6812$ training patches in total. For Biome 8 dataset, we consider $70\%$ of the patches (extracted from $32$ scenes with shadow) for training and the rest for testing to have a fair comparison against Segnet Adaption~\cite{remotesecnn8}. Generating SDAA patches for this training set and then adding them to the original training patches results in $17155$ patches in total.

After training the proposed model with training patches of the above-mentioned datasets, the obtained weights are saved and used for the evaluation of the model by prediction over unseen test scenes. Test patches of each scene are resized to $\!384\!\!\times\!\!384\!\!\times\!\!4\!$ and fed to the model. Then, the cloud/shadow probability map corresponding to each patch is generated. Next, probability maps (grayscale images) are stitched together to build a prediction map for a complete scene. A $\!50\%\!$ threshold is used to get a final  binary mask. In multiclass segmentation, an $\!argmax\!$ operation is applied on the output probability map to produce a mask for each class. Cloud prediction by Cloud-Net+ takes about $30$s for all patches of a typical Landsat 8 scene with a P100-PCIE-12GB GPU, half of which is for reading and preparing, and the other half is for inference.

\vspace{-2mm}\subsection{Evaluation Metrics}
Once cloud masks of complete Landsat 8 scenes (including empty patches) are obtained by our algorithm, they are compared against their corresponding GTs. Then, the performance is quantitatively measured by Jaccard index (or mean Intersection over Union (mIoU)), precision, recall, and accuracy. These metrics are commonly used in state-of-the-art segmentation algorithms and are defined as follows:
 
\vspace{-5mm}
\begin{equation}
\footnotesize
\begin{gathered}\hspace{-2mm}
\nonumber \text{Jaccard Index}=\frac{ \sum_{i=1}^{M} tp_{i}} {\sum_{i=1}^{M} (tp_{i} \! \!+\! \! fp_{i} \! \!+\! \! fn_{i})} ,  \text{Precision}=\frac{ \sum_{i=1}^{M} tp_{i}} {\sum_{i=1}^{M} (tp_{i} \! \!+\! \! fp_{i})} , \\
\text {Recall}=\frac{ \sum_{i=1}^{M} tp_{i}} {\sum_{i=1}^{M} (tp_{i} + fn_{i})} ,  
\text {Accuracy}=\frac{ \sum_{i=1}^{M} (tp_{i}+tn_{i})} {\sum_{i=1}^{M} (tp_{i}\! \!+\! \! tn_{i}\! \!+\! \!fp_{i}\! \!+\! \!fn_{i})} , \\
\end{gathered}
\label{Eq:Ev1}
\end{equation}
where $tp$, $tn$, $fp$, and $fn$ are the numbers of true positive, true negative, false positive, and false negative pixels for each class in each test set scene. $M$ is the total number of scenes in the test data.

\vspace{-3mm}\subsection{Quantitative and Qualitative Results}
\subsubsection{Evaluation of the Proposed Loss Functions}  
\begin{figure}[h]
\footnotesize
\begin{minipage}{0.5\textwidth}
\centering
 \includegraphics[width = 75mm]{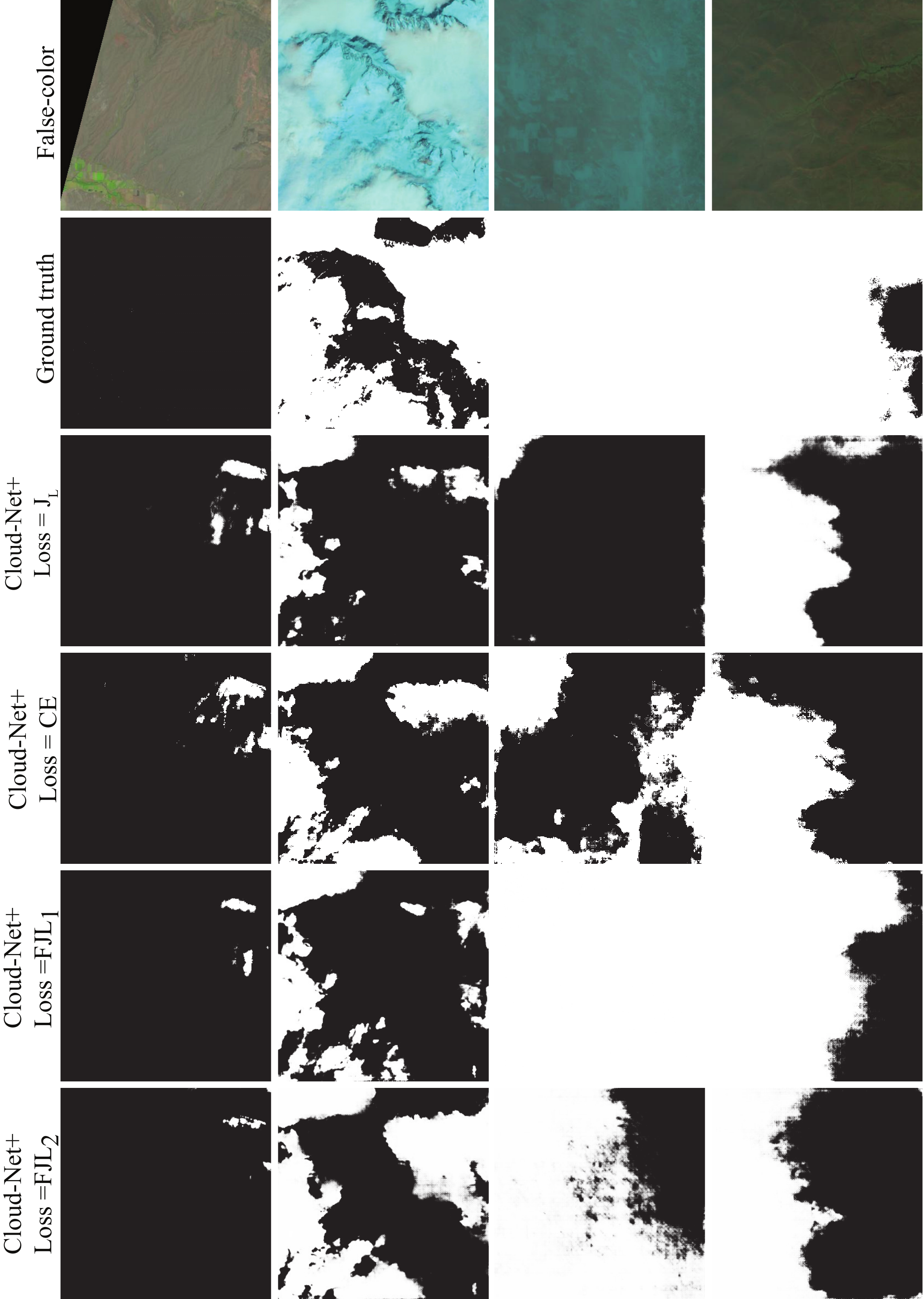}
\end{minipage}
\caption{\footnotesize Visual examples of the cloud detection results over patches of 38-Cloud test set obtained by different loss functions.  
\label{Fig:vis_losses}}
\vspace{-2mm}
\end{figure}

Table \ref{Tab:numerical_losses} demonstrates experimental results for evaluating the proposed loss functions. From Table \ref{Tab:numerical_losses}, we conclude that for both  38-Cloud and SPARCS datasets, Cloud-Net+ trained with both versions of FJL performs better than the original loss functions that they are derived from. A comparison of different loss functions over 38-Cloud dataset is displayed in Fig. \ref{Fig:vis_losses}. Note the better performance of both FJL loss functions on detecting  thick  and thin clouds (haze) over snow (two middle columns).

\renewcommand{\tabcolsep}{2pt}
\begin{table}[h]
\footnotesize 
\begin{minipage}[t]{0.49\textwidth}
\centering 
\caption{\footnotesize Quantitative results for the proposed loss functions obtained by Cloud-Net+ model for cloud detection (in~\%).
\vspace{-2mm}
\label{Tab:numerical_losses}} 
\begin{tabular}{|>{\centering\arraybackslash}m{16mm}|c|c|c|c|c|}
\hline
\centering \textbf{Training/Test}  & \textbf{Loss} & \textbf{Jaccard}                       & \textbf{Precision}   & \textbf{Recall}  & \textbf{Accuracy}   \\ \hline
\hhline{|=|=|=|=|=|=|} \hline
\multirow{4}{14mm}{\centering 38-Cloud training set /test set} & $J_{L}$ & 88.41& 97.44 & 90.51 & 96.21 \\ 
 &  proposed $F\!J\!L_1$ & \textbf{88.85} &  97.39 & \textbf{91.01}  &  \textbf{96.35} \\  
  &   $CE$ & 87.71 & 97.25 & 89.94  & 95.97 \\  
 &   proposed $F\!J\!L_2$ & 88.26 & \textbf{97.70} & 90.13  & 95.91 \\
\hhline{|=|=|=|=|=|=|} 
\multirow{4}{14mm}{\centering SPARCS folds 2-5 /fold 1}&  $J_{L}$  & 81.77 & 92.78 & 87.32 &  94.69 \\
 &  proposed $F\!J\!L_1$ & 83.99 & 92.23 & 90.37 & 95.30\\
 &  $CE$ & 81.55 & 92.25 & 87.55 & 94.60 \\
  &  proposed $F\!J\!L_2$ & \textbf{85.11} & \textbf{93.43} & \textbf{90.52} &  \textbf{95.68} \\ \hline
\end{tabular}
\end{minipage}
\vspace{0mm}
\end{table}

\vspace{-2mm}

\subsubsection{Evaluation of the Proposed SDAA Method}

Numerical results for cloud shadow detection with and without SDAA using different loss functions are reported in Table \ref{Tab:numerical_sdaa}. From this table, the addition of the augmented patches obtained by SDAA improves all evaluation metrics over SPARCS dataset compared to using original training patches only. Using SDAA to train over Biome 8 dataset also results in a higher Jaccard index and accuracy than without it.

\renewcommand{\tabcolsep}{2pt}
\vspace{-1mm}
\begin{table}[h]
\footnotesize 
\begin{minipage}[t]{0.5\textwidth}
\centering 
\caption{\footnotesize Quantitative results of shadow detection using SDAA obtained with Cloud-Net+ and different loss functions (in~\%).
\vspace{-2mm}
\label{Tab:numerical_sdaa}} 
\begin{tabular}{|>{\centering\arraybackslash}m{23mm}|c|c|c|c|c|}
\hline
\centering \textbf{Training/Test} & \textbf{Loss} & \textbf{Jaccard}                       & \textbf{Precision}   & \textbf{Recall}  & \textbf{Accuracy}   \\ \hline
\hhline{|=|=|=|=|=|=|} \hline
\multirow{4}{23mm}{\centering SPARCS folds 2-5 / fold 1} & $J_{L}$  & 58.99& 76.82 & 71.76 & 95.97 \\ 
& $CE$ & 57.55 & 79.06 & 67.90 & 95.95 \\  
& proposed $F\!J\!L_1$ & 60.59 & 75.09 & 75.84 &  96.02\\
& proposed $F\!J\!L_2$ & 60.52 & 74.91 & 75.91 & 96.00 \\\hline
\multirow{4}{23mm}{\centering SPARCS \\ folds 2-5 + SDAA patches / fold 1} &  $J_{L}$ &\textbf{62.38} & \textbf{79.05} & \textbf{74.73} & \textbf{96.36} \\ 
 &  $CE$ & \textbf{61.66} & \textbf{81.09} & \textbf{72.01} &\textbf{96.38} \\  
 & proposed $F\!J\!L_1$  & \textbf{63.44} & \textbf{79.20} & \textbf{76.12} & \textbf{96.45} \\
 &  proposed $F\!J\!L_2$ &  \textbf{63.22} & \textbf{78.41} & \textbf{76.54} & \textbf{96.40} \\ 
 \hhline{|=|=|=|=|=|=|}  
 \multirow{2}{23mm}{\centering Biome 8 \\ 70\% / 30\%} 
&  proposed $F\!J\!L_1$  & 53.79 &  58.45 & \textbf{87.09}  & 98.21  \\
& proposed $F\!J\!L_2$ & 52.11  & 56.31  & \textbf{87.47}  &  98.07 \\\hline
 \multirow{2}{23mm}{\centering Biome 8 \\ 70\%+SDAA / 30\%} 
&  proposed $F\!J\!L_1$  & \textbf{54.71}  &  \textbf{64.05} & 78.94  & \textbf{98.43}   \\ 
& proposed $F\!J\!L_2$ & \textbf{54.41} & \textbf{59.67} & 86.06  &  \textbf{98.27}   \\\hline
\end{tabular}
\end{minipage}
\end{table}

\subsubsection{Comparison with State-of-the-art Cloud Detection Methods}

Table \ref{Tab:numerical_comparison_cl} compares cloud detection results obtained by the proposed method with other methods on various datasets. From this table, the combination of the proposed Cloud-Net+ with $F\!J\!L_1$ delivers a higher Jaccard index over 38-Cloud dataset than the U-Net, Cloud-AttU, and Fmask V3 methods.  

\renewcommand{\tabcolsep}{1.5pt}
\renewcommand{\arraystretch}{1}
\begin{table}[h]
\footnotesize
\begin{minipage}[t]{0.5\textwidth}
\centering
\caption{\footnotesize Comparison of the proposed cloud detection method with other methods (in~\%).
\vspace{-2mm}
\hspace{-4mm}
\label{Tab:numerical_comparison_cl}} 
\begin{tabular}{|>{\centering\arraybackslash}m{12mm}|>{\centering\arraybackslash}m{32mm}|c|c|>{\centering\arraybackslash}m{7mm}|c|}
\hline
\centering  \textbf{Train/Test} & \textbf{Method} & \textbf{Jaccard} & \textbf{Precision}   & \textbf{Recall}  & \textbf{Accuracy}   \\ \hline
\hhline{|=|=|=|=|=|=|} \hline 
\multirow{5}{12mm}{\centering 38-Cloud training set \slash test set} & Fmask V3 \cite{fmask2}  & 85.91  & 88.65 & \textbf{96.52} &  94.94 \\  
& U-Net~\cite{mymmsp} &   85.03 & 96.15  & 88.02 &   95.05 \\ 
& Cloud-Net~\cite{myigarss} &   87.32 &  97.60 &   89.23& 95.86 \\ 
& Cloud-AttU~\cite{cloud_attu} & 88.72 &  97.16 & 91.30& \textbf{97.05} \\ 
& Cloud-Net+ + $F\!J\!L_1$ (ours) & \textbf{88.85} &  97.39 & 91.01  &  96.35 \\
& Cloud-Net+ + $F\!J\!L_2$ (ours) & 88.26 & \textbf{97.70} & 90.13  & 95.91 \\ 
\hhline{|=|=|=|=|=|=|}  
\multirow{4}{12mm}{\centering 95-Cloud training set \slash test set} & Fmask V3~\cite{fmask2}  & 85.91  & 88.65 & \textbf{96.52} & 94.94 \\ 
& Cloud-Net~\cite{myigarss} & 90.83& \textbf{97.67} & 92.84& 97.00 \\ 
& Cloud-Net+ + $F\!J\!L_1$ (ours)   &\textbf{91.57}&96.94  &94.28  & \textbf{97.23} \\ 
& Cloud-Net+ + $F\!J\!L_2$ (ours)& 91.01 & 97.49 & 93.19 & 97.06 \\
\hhline{|=|=|=|=|=|=|}  
\multirow{6}{12mm}{\centering SPARCS folds 2-5 \slash fold 1} & Fmask V3 \cite{fmask2}  & 73.01  &78.37 & \textbf{91.44} & 90.79 \\ 
& SegNet~\cite{segnet}  & 75.80  & 83.35& 89.33 & 92.23 \\
& PSPNet~\cite{pspnet}  & 78.91& 90.11& 86.39& 93.71 \\ 
& Cloud-Net~\cite{myigarss} &   81.36 &  90.81 &   88.66& 94.46 \\
& Cloud-Net+ + $F\!J\!L_1$ (ours) & 84.00 & 92.23 & 90.37 & 95.30 \\
& Cloud-Net+ + $F\!J\!L_2$ (ours) &  \textbf{85.11} & \textbf{93.43} & 90.52 & \textbf{95.68} \\ \cline{1-6} 
\multirow{3}{12mm}{\centering SPARCS 5CV} & RS-Net  \cite{rsnet} & -&  88.92 & 83.89 &  94.85 \\ 
& Cloud-Net+ + $F\!J\!L_1$ (ours)  & 83.11 & 90.75 & \textbf{90.82} & 96.20 \\
& Cloud-Net+ + $F\!J\!L_2$ (ours)  & \textbf{83.14} & \textbf{90.77} & 90.81 & \textbf{96.23} \\
\hhline{|=|=|=|=|=|=|} 
\multirow{4}{12mm}{\centering Biome 8 fold 2 \slash fold 1} &Fmask V3 \cite{fmask2}& 79.81  &82.38 & \textbf{96.24} & 93.05 \\
& Cloud-Net~\cite{myigarss} &    84.84 &  93.92 & 89.78  & 95.48 \\ 
& Cloud-Net+ + $F\!J\!L_1$ (ours) &   \textbf{85.44} & \textbf{92.88} & 91.42 & \textbf{95.55} \\
& Cloud-Net+ + $F\!J\!L_2$ (ours)  & 85.15 & 91.43 & 92.52 & 95.39 \\ \cline{1-6} 
\multirow{4}{12mm}{\centering Biome 8 2CV} & CloudFCN~\cite{cloudfcn} &-& -  &90.57& 91.00 \\
& RS-Net  \cite{rsnet} &-& \textbf{92.15}  &91.31  & 92.10  \\
& Cloud-Net+ + $F\!J\!L_1$ (ours)  & \textbf{85.15} & 90.80 & 93.28 & \textbf{95.27} \\
& Cloud-Net+ + $F\!J\!L_2$ (ours)& 83.71 & 88.69 & \textbf{93.79} & 94.67 \\ \hline
\end{tabular}
\end{minipage}
\vspace{-3mm}
\end{table}

\begin{figure}[t]
\footnotesize
\hspace{-2mm}\begin{minipage}{0.5\textwidth}
\centering
\includegraphics[width = 75mm]{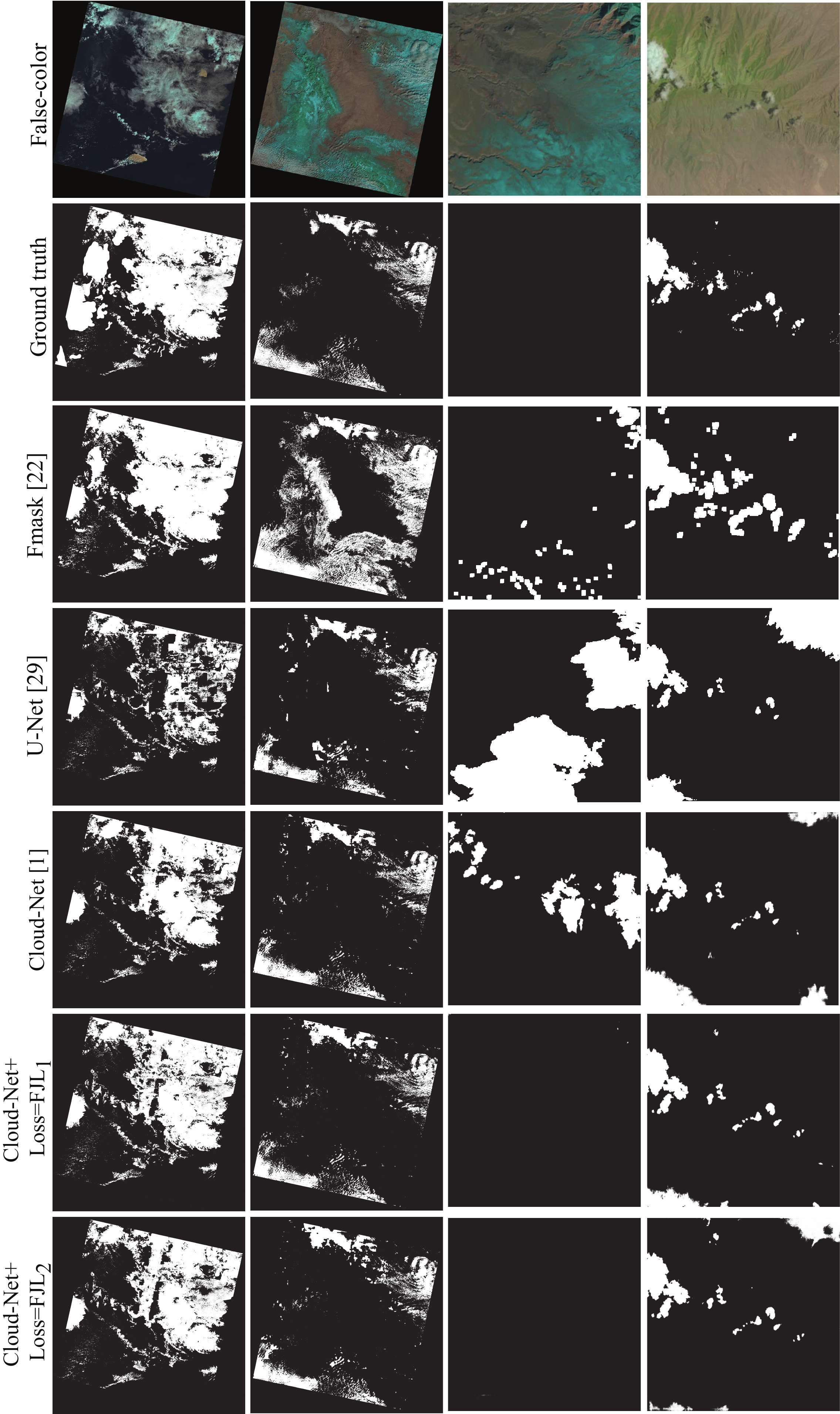}
\end{minipage}
\caption{\footnotesize Visual examples of cloud detection over 38-Cloud dataset.
\label{Fig:vis_38-cloud}}
\vspace{-3mm}
\end{figure}

Four visual examples of the predicted cloud masks are displayed in Fig. \ref{Fig:vis_38-cloud}. The second and third columns show scenes in the presence of snow. Fmask and U-Net call many $fp$s where land is covered with snow. The image of the third column does not include any clouds. The only method that successfully identifies all clear pixels is our method. The fourth column of this figure is captured over a region with bare soil---another difficult case for detecting clouds. The proposed method can distinguish those regions from clouds more accurate than the other methods.

Although it is not fair to compare our method (which only uses four spectral bands) with Fmask V4 (which uses seven spectral bands, one thermal band, DEM, and GSWO), we obtained the numerical results of Fmask V4 over 38-Cloud test set. Fmask V4 shows lower accuracy ($96.23\%$) compared to Cloud-Net+ with $F\!J\!L_1$. Note that Fmask V4 has limited applicability, as not many remote sensing platforms are equipped with more than four spectral bands of RGBNir.

\begin{figure}[h]
\footnotesize
\begin{minipage}{0.5\textwidth}
\centering
 \includegraphics[width = 80mm]{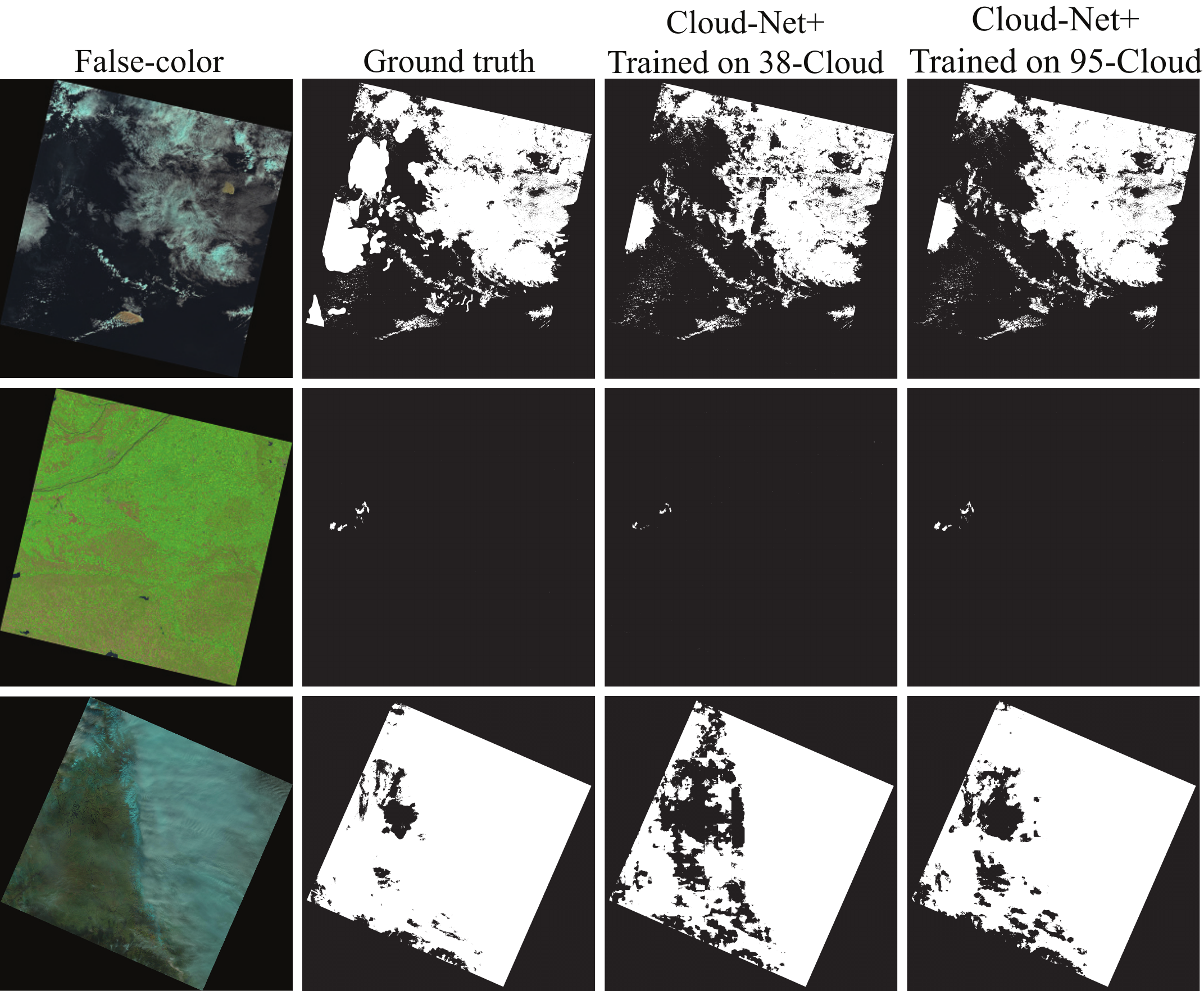}
\end{minipage}
\caption{\footnotesize Visual examples of cloud detection over 95-Cloud dataset.
\label{Fig:vis_95-cloud}}
\vspace{-3mm}
\end{figure}

The numerical results over 95-Cloud test set (which has the same test set as 38-Cloud) are improved since they are obtained by a network trained with more training images (95-Cloud training set is larger than 38-Cloud). Fig. \ref{Fig:vis_95-cloud} highlights the improvement of results over this dataset.

\begin{figure}[h]
\footnotesize
\begin{minipage}{0.5\textwidth}
\centering
 \includegraphics[width = 80mm]{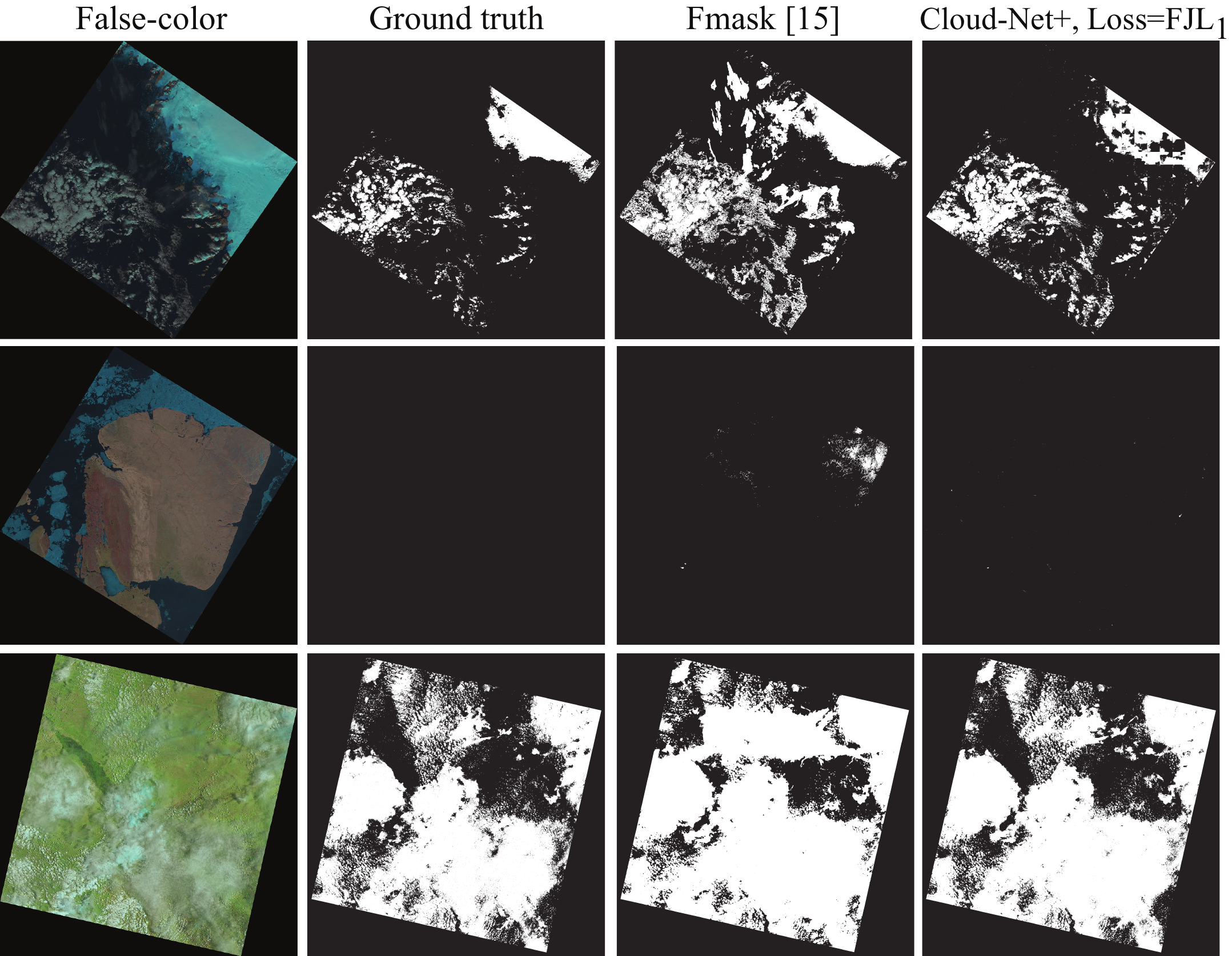}
\end{minipage}
\caption{\footnotesize Visual examples of cloud detection over Biome 8 dataset.
\label{Fig:vis_biom8}}
\vspace{-5mm}
\end{figure}

For SPARCS dataset, Cloud-Net+ outperforms SegNet~\cite{segnet} and PSPNet~\cite{pspnet}, two common semantic segmentation methods. In addition, Cloud-Net+ outperforms RS-Net over two datasets of SPARCS and Biome 8. On Biome 8 dataset, our method, which utilizes only four spectral bands, shows a higher recall and accuracy than that of CloudFCN that uses eleven spectral bands. Note that Jaccard index and accuracy take into account both  $fp$ and $fn$ in a predicted mask and penalize for all of those falsely labeled pixels, as opposed to precision and recall, which only penalize for one of the $fp$ and $fn$. That is why Jaccard index---which is not reported in some papers---and accuracy are the most important metrics when one judges different methods. Visual results over Biome 8 dataset are shown in Fig. \ref{Fig:vis_biom8}. 

\subsubsection{Comparison with State-of-the-art Cloud and Shadow Detection Methods}
\begin{figure}[h]
\footnotesize
\hspace{-2mm}\begin{minipage}{0.5\textwidth}
\centering
 \includegraphics[width = 80mm]{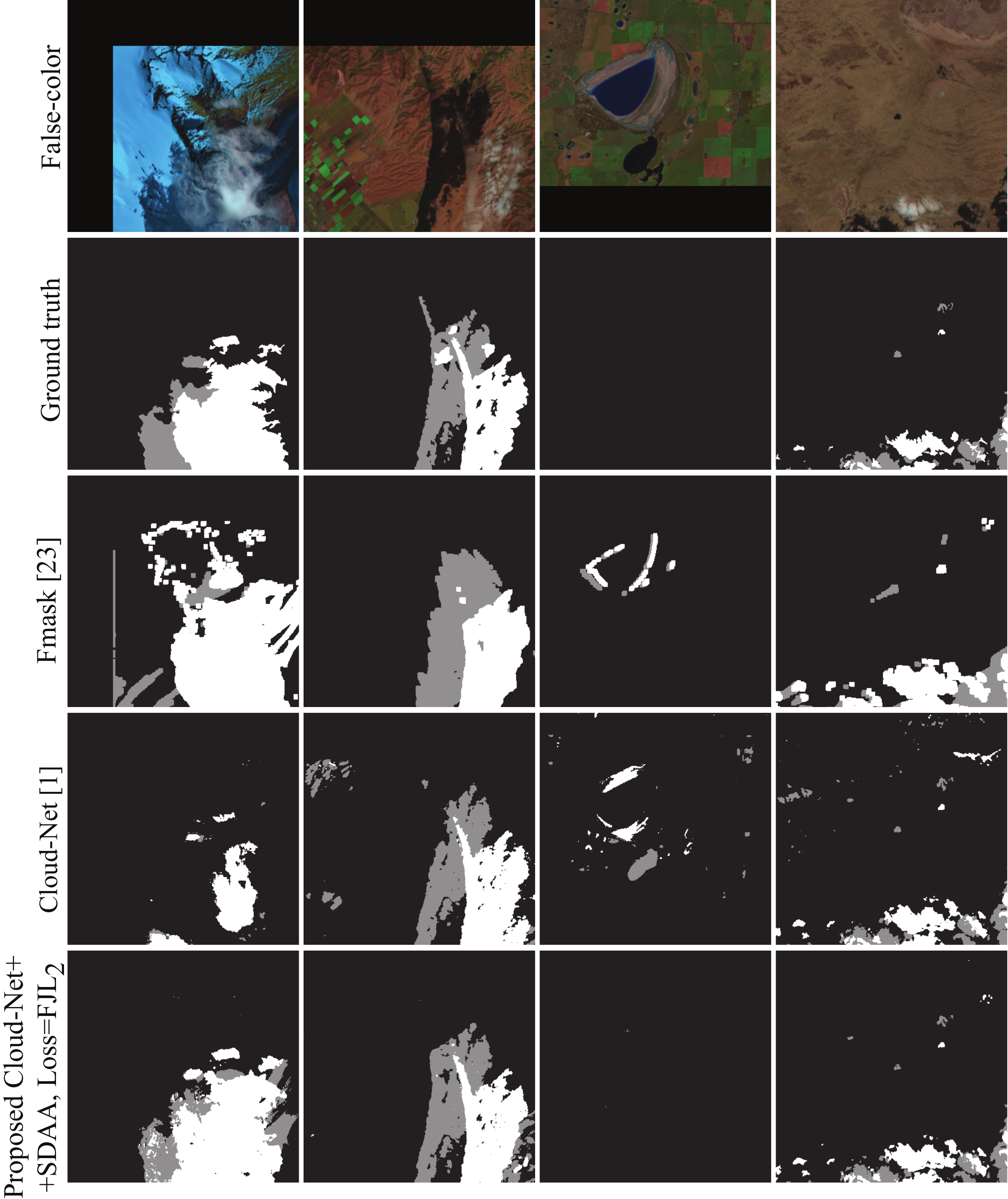}
\end{minipage}
\caption{\footnotesize Visual examples of cloud and shadow detection results over SPARCS dataset. White and gray pixels show cloud and shadow regions.
\label{Fig:vis_sparcs_clsh}}
\vspace{-3mm}
\end{figure}

The numerical results for simultaneous segmentation of cloud, shadow, and clear regions over two datasets are reported in Table \ref{Tab:numerical_comparison_shcl}.  The reported Average Jaccard in this table is the average of Jaccard indices over three classes. Two settings were considered for evaluation by the authors of SPARCS CNN: a) using 72 images from the SPARCS dataset for training and the rest of SPARCS images for testing, and b) utilizing the same 72 SPARCS images for training and 24 scenes from the Biome 8 dataset for testing (names of these 24 scenes are listed in~\cite{sparcs2}). We used the same settings in our experiments for a fair comparison. Two pixels were considered by the authors of CNN SPARCS as a leeway buffer at boundaries of predicted clouds and shadows and clear regions, where either class was considered correct within two pixels from the predicted boundary in ground truths. Note that we used no buffer pixels for the evaluation of our method. According to Table \ref{Tab:numerical_comparison_shcl}, when SPARCS CNN is trained on SPARCS dataset and evaluated on Biome 8 (even by considering a two-pixel buffer), it is outperformed by our method in terms of accuracy. However, when this method is trained and evaluated on SPARCS dataset, the numerical results are almost perfect. It should be noted that those perfect results reported for this experiment are obtained by having a two-pixel leeway buffer and using ten spectral bands, as opposed to no leeway buffer and using only four spectral bands in our method. As an example of how much this two-pixel leeway affects numerical results of the SPARCS CNN, notice the improvement in accuracy (from 91.04\% to 95.4\%), where SPARCS CNN method is trained on the SPARCS dataset and evaluated on the Biome 8. Fig. \ref{Fig:vis_sparcs_clsh}  displays some visual results for simultaneous segmentation of clouds and cloud shadows over SPARCS dataset.

For comparison againt RefUNets, we downloaded the cited the training, validation, and testing Landsat 8 scenes and conducted our experiments on them. In RefUNet v1 and RefUnet v2, the input consists of four (RGBNir) and seven spectral bands, respectively, while the default cloud and shadow masks extracted from Landsat 8 products’ QA bands are used for ground truths in experiments.  Although sophisticated boundary refinements were applied on RefUnet methods's output masks, the obtained accuracy by either of these methods is not as high as the proposed method.

\vspace{-3mm}

\renewcommand{\tabcolsep}{1.5pt}
\renewcommand{\arraystretch}{1.05}
\begin{table}[h]
\footnotesize 
\begin{minipage}[t]{0.5\textwidth}
\centering 
\caption{\footnotesize Quantitative results for simultaneous cloud and shadow detection (in~\%). In columns 2 to 4, the first row indicates metrics for the cloud class, the second row for the shadow class, and the third row for the clear class. 
\vspace{-2mm}
\label{Tab:numerical_comparison_shcl}} 
\begin{tabular}{|>{\centering\arraybackslash}m{11mm}|>{\centering\arraybackslash}m{18mm}|c|c|c|>{\centering\arraybackslash}m{11mm}|>{\centering\arraybackslash}m{12mm}|}
\hline
\centering  \textbf{Training\slash Test} & \textbf{Model} &  \textbf{Jaccard} &  \textbf{Precision} & \textbf{Recall}  & \textbf{Average Jaccard}  & \textbf{Accuracy}   \\ \hline
\hhline{|=|=|=|=|=|=|=|} \hline
\multirow{11}{11mm}{\centering SPARCS folds 2-5 /fold 1} & Fmask V3\cite{fmask2} &\begin{tabular}{@{}l@{}}  73.01 \\\tabdotline 32.24\\\tabdotline 88.41\\
                 \end{tabular} &\begin{tabular}{@{}l@{}}  78.37\\\tabdotline 52.53\\\tabdotline 95.58\\
                 \end{tabular} &\begin{tabular}{@{}l@{}}  91.44\\\tabdotline 45.51\\\tabdotline 92.18\\
                 \end{tabular} &64.56 & 92.79 \\ \cline{2-7}
& Cloud-Net\cite{myigarss}&\begin{tabular}{@{}l@{}}  82.94\\\tabdotline 60.22\\\tabdotline 93.32\\ 
                 \end{tabular} &\begin{tabular}{@{}l@{}}  90.75\\\tabdotline 76.47\\\tabdotline 96.39\\
                 \end{tabular} &\begin{tabular}{@{}l@{}}  90.60\\\tabdotline 73.92\\\tabdotline 96.70\\
                 \end{tabular} &78.83 & 96.04 \\\cline{2-7} 
& Cloud-Net+ + SDAA + $F\!J\!L_1$ (proposed)   &\begin{tabular}{@{}l@{}}  83.96\\\tabdotline 63.42\\\tabdotline 93.99\\
                 \end{tabular} &\begin{tabular}{@{}l@{}}  91.93\\\tabdotline 80.49\\\tabdotline 96.43\\
                 \end{tabular} &\begin{tabular}{@{}l@{}}  90.63\\\tabdotline 74.94\\\tabdotline 97.37\\
                 \end{tabular}  &80.45 & 96.41 \\ \cline{2-7}
& Cloud-Net+ + SDAA + $F\!J\!L_2$ (proposed)  & \begin{tabular}{@{}l@{}}  83.77\\\tabdotline 64.04\\\tabdotline 94.15\\
                 \end{tabular} &\begin{tabular}{@{}l@{}} 91.97\\\tabdotline 82.18\\\tabdotline 96.38\\
                 \end{tabular} &\begin{tabular}{@{}l@{}}  90.38\\\tabdotline 74.37\\\tabdotline 97.60\\
                 \end{tabular}   &  \textbf{80.65}  & \textbf{96.47}\\\hline 
\multirow{8}{11mm}{\centering SPARCS 72 / 8} &
SPARCS CNN~\cite{sparcs2} w/ 2-px buffer & \begin{tabular}{@{}l@{}}  92.44\\\tabdotline 87.42\\\tabdotline 97.92\\
                 \end{tabular} &\begin{tabular}{@{}l@{}} 96.42\\\tabdotline 94.87\\\tabdotline 98.75\\
                 \end{tabular} &\begin{tabular}{@{}l@{}} 95.72\\\tabdotline 91.75\\\tabdotline 99.14\\
                 \end{tabular} &  \textbf{92.59}  & \textbf{100}\\\cline{2-7}
& Cloud-Net+ + SDAA + $F\!J\!L_1$ (proposed)  & \begin{tabular}{@{}l@{}}   76.81\\\tabdotline 61.57\\\tabdotline 95.25\\
                 \end{tabular} &\begin{tabular}{@{}l@{}} 84.67\\\tabdotline 70.56\\\tabdotline 98.40\\
                 \end{tabular} &\begin{tabular}{@{}l@{}}  89.21\\\tabdotline 82.85\\\tabdotline 96.75\\
                 \end{tabular}   &  77.88  & 96.87\\\cline{2-7}
& Cloud-Net+ + SDAA + $F\!J\!L_2$ (proposed)  & \begin{tabular}{@{}l@{}}   75.56\\\tabdotline 62.32\\\tabdotline 95.29\\
                 \end{tabular} &\begin{tabular}{@{}l@{}} 85.33\\\tabdotline 71.47\\\tabdotline 98.19\\
                 \end{tabular} &\begin{tabular}{@{}l@{}}  86.83\\\tabdotline 82.97\\\tabdotline 96.99\\
                 \end{tabular}   &  77.72  & 96.87\\
\hhline{|=|=|=|=|=|=|=|}
\multirow{11}{11mm}{\centering SPARCS 72 / Biome 8 24 scenes with shadow}
& SPARCS CNN~\cite{sparcs2} w/o 2-px buffer & \begin{tabular}{@{}l@{}}  not\\ repo-\\ rted\\
                 \end{tabular} &\begin{tabular}{@{}l@{}} not\\ repo-\\ rted\\
                 \end{tabular} &\begin{tabular}{@{}l@{}} not\\ repo-\\ rted\\
                 \end{tabular}   &  not reported  & 91.04\\\cline{2-7}
& SPARCS CNN~\cite{sparcs2} w/ 2-px buffer & \begin{tabular}{@{}l@{}}  not\\ repo-\\ rted\\
                 \end{tabular} &\begin{tabular}{@{}l@{}} not\\ repo-\\ rted\\
                 \end{tabular} &\begin{tabular}{@{}l@{}} not\\ repo-\\ rted\\
                 \end{tabular}   &  not reported  & 95.40\\\cline{2-7}
& Cloud-Net+ + SDAA + $F\!J\!L_1$ (proposed)  & \begin{tabular}{@{}l@{}}   75.62\\\tabdotline 41.92\\\tabdotline 92.57\\
                 \end{tabular} &\begin{tabular}{@{}l@{}} 82.88\\\tabdotline 56.70\\\tabdotline 97.06\\
                 \end{tabular} &\begin{tabular}{@{}l@{}}  89.62\\\tabdotline 61.66\\\tabdotline 95.23\\
                 \end{tabular}   &  \textbf{70.04}  & \textbf{95.62}\\\cline{2-7}
& Cloud-Net+ + SDAA + $F\!J\!L_2$ (proposed)  & \begin{tabular}{@{}l@{}}   72.48\\\tabdotline 41.19\\\tabdotline 92.11\\
                 \end{tabular} &\begin{tabular}{@{}l@{}} 81.75\\\tabdotline 53.82\\\tabdotline 96.74\\
                 \end{tabular} &\begin{tabular}{@{}l@{}}  86.47\\\tabdotline 63.71\\\tabdotline 95.06\\
                 \end{tabular}   &  68.59  & 95.23\\
\hhline{|=|=|=|=|=|=|=|} 
\multirow{11}{11mm}{\centering Biome 8 scenes with shadow 70\% / 30\%} & Fmask V3\cite{fmask2} &\begin{tabular}{@{}l@{}}  74.88\\\tabdotline 30.51\\\tabdotline 92.51\\
                 \end{tabular} &\begin{tabular}{@{}l@{}}  75.69\\\tabdotline 41.57\\\tabdotline 99.18\
                 \end{tabular} &\begin{tabular}{@{}l@{}}  98.58\\\tabdotline 53.41\\\tabdotline 93.22\\
                 \end{tabular} &65.96 & 95.40 \\\cline{2-7}
& Segnet Adaption~\cite{remotesecnn8} &\begin{tabular}{@{}l@{}}  not\\ repo-\\ rted\\
                 \end{tabular} &\begin{tabular}{@{}l@{}}  93.20\\\tabdotline 72.20\\\tabdotline 96.88\\
                 \end{tabular} &\begin{tabular}{@{}l@{}}  93.71\\\tabdotline 71.48\\\tabdotline 96.78\\
                 \end{tabular} &not reported & 94.93 \\\cline{2-7}                  
& Cloud-Net+ + SDAA + $F\!J\!L_1$ (proposed) & \begin{tabular}{@{}l@{}}  85.53\\\tabdotline 51.20\\\tabdotline 95.26\\
                 \end{tabular} &\begin{tabular}{@{}l@{}}  93.99\\\tabdotline 55.10\\\tabdotline 98.04\\
                 \end{tabular} &\begin{tabular}{@{}l@{}}  90.47\\\tabdotline 87.85\\\tabdotline 97.10\\
                 \end{tabular}   &  77.33  & 97.29\\\cline{2-7} 
& Cloud-Net+ + SDAA + $F\!J\!L_2$ (proposed) & \begin{tabular}{@{}l@{}}  85.64\\\tabdotline 51.70\\\tabdotline 95.53\\
                 \end{tabular} &\begin{tabular}{@{}l@{}}  94.89\\\tabdotline 56.00\\\tabdotline 97.99\\
                 \end{tabular} &\begin{tabular}{@{}l@{}}  89.78\\\tabdotline 87.07\\\tabdotline 97.43\\
                 \end{tabular}   &  \textbf{77.62}  & \textbf{97.40}\\                 
\hhline{|=|=|=|=|=|=|=|}
\multirow{11}{11mm}{\centering RefUNet Landsat 8 training set / test set}
& RefUNet v1~\cite{refunet1} & \begin{tabular}{@{}l@{}}  not\\ repo-\\ rted\\
                 \end{tabular} &\begin{tabular}{@{}l@{}} 85.49\\\tabdotline 41.36\\\tabdotline 89.52\\
                 \end{tabular} &\begin{tabular}{@{}l@{}} 97.17\\\tabdotline 09.03\\\tabdotline 84.31\\
                 \end{tabular}   &  not reported  & 93.43\\\cline{2-7}
& RefUNet v2~\cite{refunet2} & \begin{tabular}{@{}l@{}}  not\\ repo-\\ rted\\
                 \end{tabular} &\begin{tabular}{@{}l@{}} 87.93\\\tabdotline 40.64\\\tabdotline 91.99\\\tabdotline
                 \end{tabular} &\begin{tabular}{@{}l@{}} 95.83\\\tabdotline 39.00\\\tabdotline 84.51\\\tabdotline
                 \end{tabular}   &  not reported  & 93.60\\\cline{2-7}
& Cloud-Net+ + SDAA + $F\!J\!L_1$ (proposed)  & \begin{tabular}{@{}l@{}}   86.52\\\tabdotline 36.48\\\tabdotline 93.06\\
                 \end{tabular} &\begin{tabular}{@{}l@{}} 94.43\\\tabdotline 58.22\\\tabdotline 95.37\\
                 \end{tabular} &\begin{tabular}{@{}l@{}}  91.17\\\tabdotline 49.42\\\tabdotline 97.46\\
                 \end{tabular}   &  \textbf{72.02}  & \textbf{95.87}\\\cline{2-7}
& Cloud-Net+ + SDAA + $F\!J\!L_2$ (proposed)  & \begin{tabular}{@{}l@{}}   86.27\\\tabdotline 36.63\\\tabdotline 92.89\\
                 \end{tabular} &\begin{tabular}{@{}l@{}} 94.59\\\tabdotline 57.01\\\tabdotline 95.28\\
                 \end{tabular} &\begin{tabular}{@{}l@{}}  90.74\\\tabdotline 50.60\\\tabdotline 97.37\\
                 \end{tabular}   &  71.93  & 95.79\\\hline
\end{tabular}
\end{minipage}
\vspace{-5mm}
\end{table}
\subsection{Recommendations on the Application of the Proposed Loss Functions}
In general, the numerical results for $F\!J\!L_1$ and $F\!J\!L_2$ are not significantly different in the majority of the experiments conducted. (see Tables \ref{Tab:numerical_comparison_cl} and \ref{Tab:numerical_comparison_shcl}). 

\begin{figure}[h]
\footnotesize
\hspace{-8mm}\begin{minipage}{.5\textwidth}
\centering
 \includegraphics[width = 100mm]{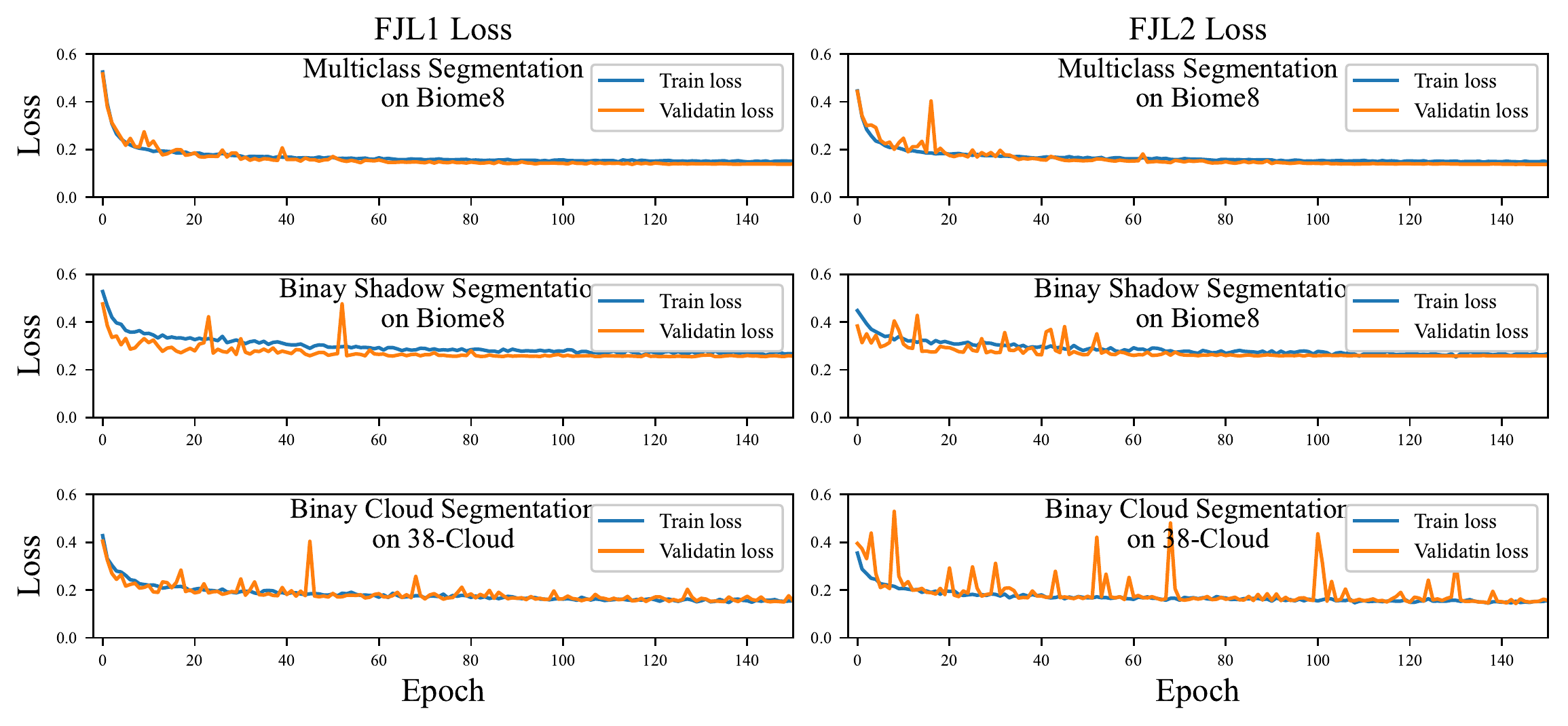}
\end{minipage}
\caption{\footnotesize Training and validation loss trend for three experiments using two versions of the proposed loss functions. For better visualization, only the first 150 epochs are shown.
\label{Fig:vis_loss_trend}}
\vspace{-3mm}
\end{figure}

Another point is that validation loss values during training seem to be more stable in $F\!J\!L_1$ than $F\!J\!L_2$, as shown in Fig. \ref{Fig:vis_loss_trend}. We believe that this is because the Inverse Jaccard Function is used as the compensatory function in $F\!J\!L_1$, which is of the same nature as the main $J_{L}$ function. However, the cross entropy, which is used as a compensatory function in $F\!J\!L_2$, is calculated very differently from $J_{L}$ as it is a log-based function. Although the range of both $F\!J\!L_1$ and $F\!J\!L_2$ is bound to [0,1], the similarity between the compensatory function and the main $J_{L}$ function in $F\!J\!L_1$ leads to a more stable validation loss trend and fewer abrupt spikes than $F\!J\!L_2$.

\vspace{-3mm}

According to Table \ref{Tab:numerical_comparison_shcl}, for multiclass segmentation of remote sensing imagery, $F\!J\!L_1$ provides a higher recall for cloud class, which means that it is more conservative in predicting non-cloud regions. As a result, for applications in which as many clear pixels as possible are required, $F\!J\!L_1$ is recommended. Otherwise, Table \ref{Tab:numerical_comparison_shcl} demonstrates very similar Jaccard and accuracy results for the identification of clouds, shadows, and clear regions in images, suggesting that either of these loss functions is effective and leads to high-quality prediction masks.

We investigated how each loss function performed over different biomes/land cover types for cloud detection and reported the results in Table \ref{Tab:numerical_biome_losses}. We used the two-fold cross validation strategy for this investigation on Biome 8 dataset, where eight different biomes are distributed equally across the two folds. According to this table, the numerical results of $F\!J\!L_1$ and $F\!J\!L_2$ are very similar in most of the biomes. The only biome type in which both Jaccard and accuracy obtained by two proposed loss functions differ by a large gap (more than 5\%), is the snow/ice biome. Table \ref{Tab:numerical_biome_losses} shows that $F\!J\!L_1$ is more effective in snowy or icy land types than $F\!J\!L_2$. As a result, we recommend using $F\!J\!L_1$ for the cloud segmentation of regions including ice, snow, and mountains, especially during the colder seasons.

\renewcommand{\tabcolsep}{1pt}
\begin{table}[h]
\footnotesize \hspace{-4mm} 
\begin{minipage}[t]{0.5\textwidth}
\centering 
\caption{\footnotesize Comparison of the cloud detection performance of the two proposed loss functions over various biome types (Jaccard/Accuracy in\%). Each reported metric is the average of the metric values obtained over fold 1 and 2.
\vspace{-1mm}
\label{Tab:numerical_biome_losses}} 
\begin{tabular}{|c|c|c|c|c|c|c|c|c|}
\hline
\multirow{2}{7mm}{\centering \textbf{Loss}} & \textbf{Snow/}&  {\centering\textbf{Shrubland}} & {\centering \textbf{Forest}}  & {\centering \textbf{Water}} &  {\centering \textbf{Urban}} &  \textbf{Grass/} & {\centering \textbf{Wetland}} &  {\centering \textbf{Barren}} \\
& \textbf{Ice} &&&&& \textbf{Crops} && \\ \hline
\hhline{|=|=|=|=|=|=|=|=|=|} \hline
\multirow{2}{7mm}{\centering $F\!J\!L_1$} & \textbf{51.8}/ & \textbf{89.7}/ &84.8/ &\textbf{91.0}/ &94.4/ &\textbf{94.2}/ &90.0/ &\textbf{87.3}/\\
& \textbf{86.1} & \textbf{96.8} &94.2 &\textbf{97.4} &\textbf{98.4} &\textbf{98.1} &96.9 &\textbf{96.5} \\\hline 
\multirow{2}{7mm}{\centering $F\!J\!L_2$} & 47.4/ & 87.9/ &\textbf{86.1}/ &90.1/ &\textbf{94.6}/ &92.1/ &\textbf{93.0}/ &87.0/\\
& 81.8 & 96.3 &\textbf{94.6} &97.2 &\textbf{98.4} &97.3 &\textbf{97.8} &96.4 \\\hline 
\end{tabular}
\end{minipage}
\end{table}

\vspace{-1mm}\subsection{Further Experiments}

To further highlight the effect of some of our choices made in this work, some experiments are conducted and summarized in Table \ref{Tab:numerical_ablation}. All experiments in this table are conveyed using soft Jaccard loss function for cloud segmentation. First, Cloud-Net+ architecture is compared against Cloud-Net (with and without using the aggregation branch, AB). As Table \ref{Tab:numerical_ablation} indicates, Jaccard index and accuracy of Cloud-Net+ are higher than those of Cloud-Net, despite Cloud-Net+'s $10\%$ less trainable parameters. Another interesting point is that by removing AB from Cloud-Net+ architecture, its performance deteriorates. This indicates that AB is capable of retrieving more details in the predicted cloud mask.

Due to computational hardware constraints, we perform our experiments with overlapping (OL) patches only on SPARCS dataset, which is the smallest of all datasets used for comparison in this work. For SPARCS, extracting and using patches with $50\%$ overlap (as suggested in ~\cite{rsnet}) improved the performance of cloud detection.

\renewcommand{\tabcolsep}{1pt}
\begin{table}[h]
\footnotesize \hspace{-3mm} 
\begin{minipage}[t]{0.5\textwidth}
\centering 
\caption{\footnotesize Quantitative results for further experiments (in~\%).
\vspace{-2mm}
\label{Tab:numerical_ablation}} 
\begin{tabular}{|>{\centering\arraybackslash}m{17mm}|c|c|c|c|c|}
\hline
\centering \textbf{\textbf{Training/Test}} & \textbf{Model} &  \textbf{Jaccard}                       & \textbf{Precision}   & \textbf{Recall}  & \textbf{Accuracy}   \\ \hline
\hhline{|=|=|=|=|=|=|} \hline
\multirow{3}{17mm}{\centering 38-Cloud training set /test set} &  Cloud-Net  &   87.32 &  97.60 &  89.23& 95.86 \\ \cline{2-6}
 & Cloud-Net+ w/o AB & 87.84&  \textbf{97.84} & 89.57  &  96.04 \\ \cline{2-6}
 & Cloud-Net+ &  \textbf{88.41}&  97.44 & \textbf{90.51}  &  \textbf{96.21} \\ \cline{2-6}
\hhline{|=|=|=|=|=|=|}  
\multirow{2}{17mm}{\centering SPARCS folds 2-5 \slash fold 1} & Cloud-Net+ w/o OL  & 81.77 & \textbf{92.78} & 87.32 &  94.69  \\ \cline{2-6}
 & Cloud-Net+ w OL  & \textbf{83.35} &  92.08 &  \textbf{ 89.78}& \textbf{95.11} \\ \hline
\end{tabular}
\end{minipage}
\vspace{-3mm}
\end{table}

\vspace{-3mm}\subsection{Experiment on Pascal VOC Dataset}
To test the proposed Filtered Jaccard Loss functions beyond cloud/shadow detection applications, we have conducted experiments over Pascal VOC 2012 semantic segmentation dataset\cite{pascal_dataset, pascal_dataset_aug}. This dataset contains $10582$ images for training and $1449$ for testing. Each pixel in each image has been assigned to one of the  $21$ existing classes, including airplane, bicycle, cat, horse, and person, to name a few. To obtain a binary segmentation, we considered pixels of only one of the classes---person class---as the positive class and all other pixels in each image as the negative class.
Numerical results in Table \ref{Tab:pascalvoc_numerical} indicate that both versions of the proposed loss function outperform soft Jaccard loss for other types of images in addition to the remote sensing ones.

\renewcommand{\tabcolsep}{4pt}
\vspace{-2mm}\begin{table}[h]
\footnotesize
\begin{minipage}[t]{0.47\textwidth}
\centering
\caption{\footnotesize Quantitative results over Pascal VOC dataset (in~\%).
\vspace{-2mm}
\label{Tab:pascalvoc_numerical}} 
\begin{tabular}{|c|c|c|c|}
\hline
\centering \textbf{Method} & \textbf{Loss} & \textbf{Jaccard}&  \textbf{Accuracy}         \\ \hline
\hhline{|=|=|=|=|} \hline
Cloud-Net+ &  $J_{L}$ & 58.63 & 97.19\\ \hline 
Cloud-Net+ &  $F\!J\!L_1$ & \textbf{63.11} & \textbf{97.59}\\ \hline 
Cloud-Net+ &  $F\!J\!L_2$ & 62.76 & 97.37\\ \hline
\end{tabular}
\end{minipage}

\end{table}

\section{Conclusion}
In this work, we have developed a novel system for cloud and cloud shadow detection in Landsat 8 imagery using a deep learning-based approach. The proposed network, Cloud-Net+, benefits from multiple convolution blocks that extract global and local cloud/shadow features. This network, which outperforms its parent network, Cloud-Net, has been optimized using two new and novel loss functions that reduce the number of misclassified pixels. These loss functions can be used for other binary or multiclass segmentation tasks, where the target object exists only in some of the images. In addition, a new augmentation technique, Sunlight Direction-Aware Augmentation (SDAA), is introduced for the task of cloud shadow detection. SDAA takes into account solar angles to generate natural-looking synthetic RGBNir images. We have also released an extension to our previously introduced cloud detection dataset. It will help researchers to improve the generalization ability of their  cloud segmentation algorithms.

\section*{Acknowledgment}
The authors would like to thank the Government of Canada
and MacDonald, Dettwiler and Associates Ltd. for the financial
support for this research through the Technology
Development Program. Also, this research was enabled in part
by the technical support provided by Compute Canada.



\vspace{-1mm}
{\footnotesize
\bibliographystyle{IEEEtran}
\bibliography{refs}}

\begin{thebibliography}{10}
\providecommand{\url}[1]{#1}
\csname url@samestyle\endcsname
\providecommand{\newblock}{\relax}
\providecommand{\bibinfo}[2]{#2}
\providecommand{\BIBentrySTDinterwordspacing}{\spaceskip=0pt\relax}
\providecommand{\BIBentryALTinterwordstretchfactor}{4}
\providecommand{\BIBentryALTinterwordspacing}{\spaceskip=\fontdimen2\font plus
\BIBentryALTinterwordstretchfactor\fontdimen3\font minus
  \fontdimen4\font\relax}
\providecommand{\BIBforeignlanguage}[2]{{%
\expandafter\ifx\csname l@#1\endcsname\relax
\typeout{** WARNING: IEEEtran.bst: No hyphenation pattern has been}%
\typeout{** loaded for the language `#1'. Using the pattern for}%
\typeout{** the default language instead.}%
\else
\language=\csname l@#1\endcsname
\fi
#2}}
\providecommand{\BIBdecl}{\relax}
\BIBdecl

\bibitem{myigarss}
S.~{Mohajerani} and P.~{Saeedi}, ``{Cloud-Net: An End-to-End Cloud Detection
  Algorithm for Landsat 8 Imagery},'' in \emph{IEEE Int. Geos. and Rem. Sens.
  Symp.}, 2019, pp. 1029--1032.

\bibitem{cloud_free_sky}
M.~D. {King}, S.~{Platnick}, W.~P. {Menzel}, S.~A. {Ackerman}, and P.~A.
  {Hubanks}, ``{Spatial and Temporal Distribution of Clouds Observed by MODIS
  Onboard the Terra and Aqua Satellites},'' \emph{IEEE Trans. on Geosc. and
  Rem. Sens.}, vol.~51, no.~7, pp. 3826--3852, 2013.

\bibitem{cloud_morph}
C.~R. {Mirza}, T.~{Koike}, K.~{Yang}, and T.~{Graf}, ``{The Development of 1-D
  Ice Cloud Microphysics Data Assimilation System (IMDAS) for Cloud Parameter
  Retrievals by Integrating Satellite Data},'' in \emph{IEEE Int. Geosc. and
  Rem. Sens. Simp.}, vol.~2, 2008, pp. II--501--II--504.

\bibitem{volcano}
L.~Zhu, M.~Wang, J.~Shao, C.~Liu, C.~Zhao, and Y.~Zhao, ``{Remote Sensing of
  Global Volcanic Eruptions Using Fengyun Series Satellites},'' in \emph{IEEE
  Int. Geosc. and Rem. Sens. Simp.}, 2015, pp. 4797--4800.

\bibitem{hurricane}
R.~S. Reddy, D.~Lu, F.~Tuluri, and M.~Fadavi, ``{Simulation and Prediction of
  Hurricane Lili During Landfall over the Central Gulf States Using MM5
  Modeling System and Satellite Data},'' in \emph{IEEE Int. Geosc. and Rem.
  Sens. Simp.}, 2017, pp. 36--39.

\bibitem{all_chinese_sats}
M.~{Huang}, X.~{Xing}, W.~{Luo}, and Z.~{Wang}, ``Recalibration of offshore
  chlorophyll content based on virtual satellite constellation,'' in \emph{IEEE
  Int. Geosc. and Rem. Sens. Simp.}, 2019, pp. 7976--7979.

\bibitem{thresholdbased8}
H.~Zhai, H.~Zhang, L.~Zhang, and P.~Li, ``Cloud/shadow detection based on
  spectral indices for multi/hyperspectral optical remote sensing imagery,''
  \emph{ISPRS J. of Photogram. and Rem. Sen.}, vol. 144, pp. 235 -- 253, 2018.

\bibitem{thresholdbased9}
Z.~Li, H.~Shen, H.~Li, G.~Xia, P.~Gamba, and L.~Zhang, ``{Multi-feature
  combined cloud and cloud shadow detection in GaoFen-1 wide field of view
  imagery},'' \emph{Rem. Sens. of Env.}, vol. 191, pp. 342 -- 358, 2017.

\bibitem{thresholdbased11}
S.~Qiu, Z.~Zhu, and C.~E. Woodcock, ``Cirrus clouds that adversely affect
  landsat 8 images: What are they and how to detect them?'' \emph{Rem. Sens. of
  Env.}, vol. 246, p. 111884, 2020.

\bibitem{thresholdbased13}
X.~Zhu and E.~H. Helmer, ``An automatic method for screening clouds and cloud
  shadows in optical satellite image time series in cloudy regions,''
  \emph{Rem. Sens. of Env.}, vol. 214, pp. 135 -- 153, 2018.

\bibitem{hot}
Y.~{Zhang}, B.~{Guindon}, and J.~{Cihlar}, ``{An Image Transform to
  Characterize and Compensate for Spatial Variations in Thin Cloud
  Contamination of Landsat Images},'' \emph{Rem. Sen. of Env.}, vol.~82, no.~2,
  pp. 173 -- 187, 2002.

\bibitem{handcraft11}
Z.~{An} and Z.~{Shi}, ``Scene learning for cloud detection on remote-sensing
  images,'' \emph{IEEE J. of Selec. Top. in Appl. Earth Obs. and Rem. Sen.},
  vol.~8, no.~8, pp. 4206--4222, 2015.

\bibitem{handcraft12}
R.~{Luo}, W.~{Liao}, H.~{Zhang}, L.~{Zhang}, P.~{Scheunders}, Y.~{Pi}, and
  W.~{Philips}, ``Fusion of hyperspectral and lidar data for classification of
  cloud-shadow mixed remote sensed scene,'' \emph{IEEE J. of Selec. Top. in
  Appl. Earth Obs. and Rem. Sen.}, vol.~10, no.~8, pp. 3768--3781, 2017.

\bibitem{handcraft13}
P.~{Bo}, S.~{Fenzhen}, and M.~{Yunshan}, ``{A Cloud and Cloud Shadow Detection
  Method Based on Fuzzy c-Means Algorithm},'' \emph{IEEE J. of Selec. Top. in
  Appl. Earth Obs. and Rem. Sen.}, vol.~13, pp. 1714--1727, 2020.

\bibitem{remotesecnn4}
S.~{Ji}, P.~{Dai}, M.~{Lu}, and Y.~{Zhang}, ``Simultaneous cloud detection and
  removal from bitemporal remote sensing images using cascade convolutional
  neural networks,'' \emph{IEEE Trans. on Geosc. and Rem. Sens.}, pp. 1--17,
  2020.

\bibitem{remotesecnn5}
M.~Segal-Rozenhaimer, A.~Li, K.~Das, and V.~Chirayath, ``{Cloud detection
  algorithm for multi-modal satellite imagery using convolutional
  neural-networks (CNN)},'' \emph{Rem. Sens. of Env.}, vol. 237, p. 111446,
  2020.

\bibitem{remotesecnn6}
Z.~Li, H.~Shen, Q.~Cheng, Y.~Liu, S.~You, and Z.~He, ``Deep learning based
  cloud detection for medium and high resolution remote sensing images of
  different sensors,'' \emph{ISPRS J. of Photogram. and Rem. Sen.}, vol. 150,
  pp. 197 -- 212, 2019.

\bibitem{remotesecnn7}
G.~{Mateo-Garc\'{i}a}, V.~Laparra, D.~López-Puigdollers, and
  L.~{G\'{o}mez-Chova}, ``Transferring deep learning models for cloud detection
  between landsat-8 and proba-v,'' \emph{ISPRS J. of Photogram. and Rem. Sen.},
  vol. 160, pp. 1 -- 17, 2020.

\bibitem{multilevel}
F.~Xie, M.~Shi, Z.~Shi, J.~Yin, and D.~Zhao, ``{Multilevel Cloud Detection in
  Remote Sensing Images Based on Deep Learning},'' \emph{IEEE J. of Selec. Top.
  in Appl. Earth Obs. and Rem. Sens.}, vol.~10, no.~8, pp. 3631--3640, 2017.

\bibitem{remotesecnn12}
Y.~Li, W.~Chen, Y.~Zhang, C.~Tao, R.~Xiao, and Y.~Tan, ``{Accurate cloud
  detection in high-resolution remote sensing imagery by weakly supervised deep
  learning},'' \emph{Rem. Sens. of Env.}, vol. 250, p. 112045, 2020.

\bibitem{fmask1}
Z.~Zhu and C.~E. Woodcock, ``{Object-Based Cloud and Cloud Shadow Detection in
  Landsat Imagery},'' \emph{Rem. Sen. of Env.}, vol. 118, pp. 83 -- 94, 2012.

\bibitem{fmask2}
Z.~Zhu, S.~Wang, and C.~E. Woodcock, ``{Improvement and Expansion of the Fmask
  Algorithm: Cloud, Cloud Shadow, and Snow Detection for Landsats 4–7, 8, and
  Sentinel 2 Images},'' \emph{Rem. Sen. of Env.}, vol. 159, pp. 269 -- 277,
  2015.

\bibitem{fmask4}
S.~Qiu, Z.~Zhu, and B.~He, ``{Fmask 4.0: Improved Cloud and Cloud Shadow
  Detection in Landsats 4\textendash 8 and Sentinel-2 Imagery},'' \emph{Rem.
  Sen. of Env.}, vol. 231, p. 111205, 2019.

\bibitem{acca}
R.~R.~Irish, J.~L.~Barker, S.~Goward, and T.~Arvidson, ``{Characterization of
  the Landsat-7 ETM+ Automated Cloud-Cover Assessment (ACCA) Algorithm},''
  \emph{Photogram. Eng. and Rem. Sen.}, vol.~72, pp. 1179--1188, 2006.

\bibitem{hot2}
Y.~{Zhang}, B.~{Guindon}, and X.~{Li}, ``{A Robust Approach for Object-Based
  Detection and Radiometric Characterization of Cloud Shadow Using Haze
  Optimized Transformation},'' \emph{IEEE Trans. on Geosc. and Rem. Sens.},
  vol.~52, no.~9, pp. 5540--5547, 2014.

\bibitem{hot3}
S.~{Chen}, X.~{Chen}, J.~{Chen}, P.~{Jia}, X.~{Cao}, and C.~{Liu}, ``An
  iterative haze optimized transformation for automatic cloud/haze detection of
  landsat imagery,'' \emph{IEEE Trans. on Geosc. and Rem. Sens.}, vol.~54,
  no.~5, pp. 2682--2694, 2016.

\bibitem{temporal4}
L.~{Xu}, A.~{Wong}, and D.~A. {Clausi}, ``A novel bayesian spatial–temporal
  random field model applied to cloud detection from remotely sensed imagery,''
  \emph{IEEE Trans. on Geosc. and Rem. Sens.}, vol.~55, no.~9, pp. 4913--4924,
  2017.

\bibitem{remotesecnn11}
A.~Francis, P.~Sidiropoulos, and J.~Muller, ``{CloudFCN: Accurate and Robust
  Cloud Detection for Satellite Imagery with Deep Learning},'' \emph{Rem.
  Sen.}, vol.~11, no.~19, 2019.

\bibitem{mymmsp}
S.~Mohajerani, T.~A. Krammer, and P.~Saeedi, ``{A Cloud Detection Algorithm for
  Remote Sensing Images Using Fully Convolutional Neural Networks},'' in
  \emph{IEEE Int. Workshop on Multim. Sig. Proc.}, 2018, pp. 1--5.

\bibitem{unet}
O.~Ronneberger, P.~Fischer, and T.~Brox, ``{U-Net: Convolutional Networks for
  Biomedical Image Segmentation},'' \emph{CoRR}, vol. abs/1505.04597, 2015.

\bibitem{shadow_1}
B.~{Zhong}, W.~{Chen}, S.~{Wu}, L.~{Hu}, X.~{Luo}, and Q.~{Liu}, ``{A Cloud
  Detection Method Based on Relationship Between Objects of Cloud and
  Cloud-Shadow for Chinese Moderate to High Resolution Satellite Imagery},''
  \emph{IEEE J. of Selected Topics in Applied Earth Observations and Remote
  Sensing}, vol.~10, no.~11, pp. 4898--4908, 2017.

\bibitem{handcraft10}
X.~{Zhang}, L.~{Liu}, X.~{Chen}, S.~{Xie}, and L.~{Lei}, ``{A Novel
  Multitemporal Cloud and Cloud Shadow Detection Method Using the Integrated
  Cloud Z-Scores Model},'' \emph{IEEE J. of Selec. Top. in Appl. Earth Obs. and
  Rem. Sen.}, vol.~12, no.~1, pp. 123--134, 2019.

\bibitem{gonzalo_multitemporal}
G.~Mateo-Garc\'{i}a, L.~G\'{o}mez-Chova, J.~Amorós-López, J.~Muñoz-Marí,
  and G.~Camps-Valls, ``{Multitemporal Cloud Masking in the Google Earth
  Engine},'' \emph{Rem. Sen.}, vol.~10, no.~7, 2018.

\bibitem{first_cloud_classification}
J.~{Kittler} and D.~{Pairman}, ``{Contextual Pattern Recognition Applied to
  Cloud Detection and Identification},'' \emph{IEEE Trans. on Geosc. and Rem.
  Sens.}, vol. GE-23, no.~6, pp. 855--863, 1985.

\bibitem{temporal3}
G.~{Vivone}, P.~{Addesso}, R.~{Conte}, M.~{Longo}, and R.~{Restaino}, ``A class
  of cloud detection algorithms based on a map-mrf approach in space and
  time,'' \emph{IEEE Trans. on Geosc. and Rem. Sens.}, vol.~52, no.~8, pp.
  5100--5115, 2014.

\bibitem{temporal6}
D.~S. Candra, S.~Phinn, and P.~Scarth, ``{Automated Cloud and Cloud-Shadow
  Masking for Landsat 8 Using Multitemporal Images in a Variety of
  Environments},'' \emph{Rem. Sen.}, vol.~11, no.~17, 2019.

\bibitem{temporal7}
S.~{Ji}, P.~{Dai}, M.~{Lu}, and Y.~{Zhang}, ``{Simultaneous Cloud Detection and
  Removal From Bitemporal Remote Sensing Images Using Cascade Convolutional
  Neural Networks},'' \emph{IEEE Trans. on Geosc. and Rem. Sens.}, pp. 1--17,
  2020.

\bibitem{pcanet}
Y.~Zi, F.~Xie, and Z.~Jiang, ``{A Cloud Detection Method for Landsat 8 Images
  Based on PCANet},'' \emph{Rem. Sen.}, vol.~10, no.~6, 2018.

\bibitem{remotesecnn9}
N.~Chen, W.~Li, C.~Gatebe, T.~Tanikawa, M.~Hori, R.~Shimada, T.~Aoki, and
  K.~Stamnes, ``New neural network cloud mask algorithm based on radiative
  transfer simulations,'' \emph{Rem. Sens. of Env.}, vol. 219, pp. 62 -- 71,
  2018.

\bibitem{remotesecnn1}
Z.~{Shao}, Y.~{Pan}, C.~{Diao}, and J.~{Cai}, ``{Cloud Detection in Remote
  Sensing Images Based on Multiscale Features-Convolutional Neural Network},''
  \emph{IEEE Trans. on Geosc. and Rem. Sens.}, vol.~57, no.~6, pp. 4062--4076,
  2019.

\bibitem{remotesecnn2}
W.~Xie, J.~Yang, Y.~Li, J.~Lei, J.~Zhong, and J.~Li, ``{Discriminative Feature
  Learning Constrained Unsupervised Network for Cloud Detection in Remote
  Sensing Imagery},'' \emph{Rem. Sen.}, vol.~12, no.~3, 2020.

\bibitem{remotesecnn3}
J.~{Lu}, Y.~{Wang}, Y.~{Zhu}, X.~{Ji}, T.~{Xing}, W.~{Li}, and A.~Y. {Zomaya},
  ``{PSegnet and NPSegnet: New Neural Network Architectures for Cloud
  Recognition of Remote Sensing Images},'' \emph{IEEE Access}, vol.~7, pp.
  87\,323--87\,333, 2019.

\bibitem{remotesecnn10}
M.~Wieland, Y.~Li, and S.~Martinis, ``{Multi-sensor cloud and cloud shadow
  segmentation with a convolutional neural network},'' \emph{Rem. Sens. of
  Env.}, vol. 230, p. 111203, 2019.

\bibitem{cdnet}
J.~{Yang}, J.~{Guo}, H.~{Yue}, Z.~{Liu}, H.~{Hu}, and K.~{Li}, ``{CDnet:
  CNN-Based Cloud Detection for Remote Sensing Imagery},'' \emph{IEEE Trans. on
  Geosc. and Rem. Sens.}, pp. 1--17, 2019.

\bibitem{cdnet2}
J.~{Guo}, J.~{Yang}, H.~{Yue}, H.~{Tan}, C.~{Hou}, and K.~{Li}, ``{CDnetV2:
  CNN-Based Cloud Detection for Remote Sensing Imagery With Cloud-Snow
  Coexistence},'' \emph{IEEE . on Geosc. and Rem Sens}, pp. 1--14, 2020.

\bibitem{remotesecnn8}
D.~Chai, S.~Newsam, H.~K. Zhang, Y.~Qiu, and J.~Huang, ``Cloud and cloud shadow
  detection in landsat imagery based on deep convolutional neural networks,''
  \emph{Rem. Sens. of Env.}, vol. 225, pp. 307 -- 316, 2019.

\bibitem{rsnet}
J.~H. Jeppesen, R.~H. Jacobsen, F.~Inceoglu, and T.~S. Toftegaard, ``{A Cloud
  Detection Algorithm for Satellite Imagery Based on Deep Learning},''
  \emph{Rem. Sen. of Env.}, vol. 229, pp. 247 -- 259, 2019.

\bibitem{sparcs2}
M.~J. Hughes and R.~Kennedy, ``{High-Quality Cloud Masking of Landsat 8 Imagery
  Using Convolutional Neural Networks},'' \emph{Rem. Sen.}, vol.~11, no.~21,
  2019.

\bibitem{refunet1}
L.~Jiao, L.~Huo, C.~Hu, and P.~Tang, ``{Refined UNet: UNet-Based Refinement
  Network for Cloud and Shadow Precise Segmentation},'' \emph{Remote Sensing},
  vol.~12, no.~12, 2020.

\bibitem{refunet2}
------, ``Refined unet v2: End-to-end patch-wise network for noise-free cloud
  and shadow segmentation,'' \emph{Remote Sensing}, vol.~12, no.~21, 2020.

\bibitem{cloud_attu}
Y.~Guo, X.~Cao, B.~Liu, and M.~Gao, ``{Cloud Detection for Satellite Imagery
  Using Attention-Based U-Net Convolutional Neural Network},'' \emph{Symmetry},
  vol.~12, no.~6, 2020.

\bibitem{cloudfcn}
A.~Francis, P.~Sidiropoulos, and J.~Muller, ``{CloudFCN: Accurate and Robust
  Cloud Detection for Satellite Imagery with Deep Learning},'' \emph{Remote
  Sensing}, vol.~11, no.~19, 2019.

\bibitem{aug_gan1}
D.~{Ma}, P.~{Tang}, and L.~{Zhao}, ``{SiftingGAN: Generating and Sifting
  Labeled Samples to Improve the Remote Sensing Image Scene Classification
  Baseline In Vitro},'' \emph{IEEE Geosc. and Rem. Sens. Letters}, vol.~16,
  no.~7, pp. 1046--1050, 2019.

\bibitem{aug_gan2}
J.~Howe, K.~Pula, and A.~A. Reite, ``{Conditional generative adversarial
  networks for data augmentation and adaptation in remotely sensed imagery},''
  in \emph{Applications of Mach. Learn.}, M.~E. Zelinski, T.~M. Taha, J.~Howe,
  A.~A.~S. Awwal, and K.~M. Iftekharuddin, Eds., vol. 11139.\hskip 1em plus
  0.5em minus 0.4em\relax SPIE, 2019, pp. 119--131.

\bibitem{aug_gan3}
K.~Zheng, M.~Wei, G.~Sun, B.~Anas, and Y.~Li, ``{Using Vehicle Synthesis
  Generative Adversarial Networks to Improve Vehicle Detection in Remote
  Sensing Images},'' \emph{ISPRS Int. J. of Geo-Info.}, vol.~8, no.~9, 2019.

\bibitem{shadow_2}
D.~Marmanis, K.~Schindler, J.~Wegner, S.~Galliani, M.~Datcu, and U.~Stilla,
  ``{Classification with an edge: Improving semantic image segmentation with
  boundary detection},'' \emph{ISPRS J. of Photogrammetry and Remote Sensing},
  vol. 135, pp. 158 -- 172, 2018.

\bibitem{myicip2019}
S.~{Mohajerani}, R.~{Asad}, K.~{Abhishek}, N.~{Sharma}, A.~v.~{Duynhoven}, and
  P.~{Saeedi}, ``{CloudMaskGAN: A Content-Aware Unpaired Image-to-image
  Translation Algorithm for Remote Sensing Imagery},'' in \emph{IEEE Int. Conf.
  on Image Proc.}, 2019, pp. 1965--1969.

\bibitem{net_in_net}
M.~Lin, Q.~Chen, and S.~Yan, ``{Network In Network},'' \emph{CoRR}, vol.
  abs/1312.4400, 2013.

\bibitem{wide-res}
S.~Zagoruyko and N.~Komodakis, ``{Wide Residual Networks},'' in
  \emph{Proceedings of the British Mach. Vision Conf.}, E.~R.~H. Richard
  C.~Wilson and W.~A.~P. Smith, Eds., 2016, pp. 87.1--87.12.

\bibitem{jacc1}
W.~Waegeman, K.~Dembczy\'{n}ki, A.~Jachnik, W.~Cheng, and E.~H\"{u}llermeier,
  ``{On the Bayes-optimality of F-measure Maximizers},'' \emph{J. of Mach.
  Learn. Research}, vol.~15, no.~1, pp. 3333--3388, 2014.

\bibitem{dice1}
A.~A. {Novikov}, D.~{Lenis}, D.~{Major}, J.~{Hladůvka}, M.~{Wimmer}, and
  K.~{Bühler}, ``{Fully Convolutional Architectures for Multiclass
  Segmentation in Chest Radiographs},'' \emph{IEEE TMI}, vol.~37, no.~8, pp.
  1865--1876, 2018.

\bibitem{jacc2}
Y.~Yuan, M.~Chao, and Y.~C. Lo, ``{Automatic Skin Lesion Segmentation Using
  Deep Fully Convolutional Networks With Jaccard Distance},'' \emph{IEEE Trans.
  on Medical Imaging}, vol.~36, no.~9, pp. 1876--1886, 2017.

\bibitem{loss_idea}
X.~Lu, C.~Ma, B.~Ni, X.~Yang, I.~Reid, and M.~Yang, ``{Deep Regression Tracking
  with Shrinkage Loss},'' in \emph{European Conf. on Comp. Vision}, 2018.

\bibitem{ADAM}
D.~P. Kingma and J.~Ba, ``{Adam: {A} Method for Stochastic Optimization},''
  \emph{CoRR}, vol. abs/1412.6980, 2014.

\bibitem{xavier}
X.~Glorot and Y.~Bengio, ``{"Understanding the difficulty of training deep
  feedforward neural networks"},'' in \emph{Proc. of Int. Conf. on Artificial
  Intelligence and Statistics}, ser. Proc. of Mach. Learn. Research, vol.~9,
  2010, pp. 249--256.

\bibitem{biome8_1}
S.~Foga, P.~L. Scaramuzza, S.~Guo, Z.~Zhu, R.~D. Dilley, T.~Beckmann, G.~L.
  Schmidt, J.~L. Dwyer, M.~J. Hughes, and B.~Laue, ``{Cloud Detection Algorithm
  Comparison and Validation for Operational {L}andsat Data Products},''
  \emph{Rem. Sen. of Env.}, vol. 194, pp. 379 -- 390, 2017.

\bibitem{sparcs}
M.~J. Hughes and D.~J. Hayes, ``{Automated Detection of Cloud and Cloud Shadow
  in Single-Date Landsat Imagery Using Neural Networks and Spatial
  Post-Processing},'' \emph{Rem. Sen.}, vol.~6, no.~6, pp. 4907--4926, 2014.

\bibitem{segnet}
V.~{Badrinarayanan}, A.~{Kendall}, and R.~{Cipolla}, ``{SegNet: A Deep
  Convolutional Encoder-Decoder Architecture for Image Segmentation},''
  \emph{IEEE TPAMI}, vol.~39, no.~12, pp. 2481--2495, 2017.

\bibitem{pspnet}
H.~{Zhao}, J.~{Shi}, X.~{Qi}, X.~{Wang}, and J.~{Jia}, ``{Pyramid Scene Parsing
  Network},'' in \emph{IEEE Conf. on Comp. Vision and Patt. Recog.}, 2017, pp.
  6230--6239.

\bibitem{pascal_dataset}
M.~Everingham, L.~Van~Gool, C.~K.~I. Williams, J.~Winn, and A.~Zisserman,
  ``{The {PASCAL} {V}isual {O}bject {C}lasses {C}hallenge 2012 {(VOC2012)}
  {R}esults},''
  http://www.pascal-network.org/challenges/VOC/voc2012/workshop/index.html.

\bibitem{pascal_dataset_aug}
B.~{Hariharan}, P.~{Arbeláez}, L.~{Bourdev}, S.~{Maji}, and J.~{Malik},
  ``Semantic contours from inverse detectors,'' in \emph{Int. Conf. on Comp.
  Vision}, 2011, pp. 991--998.

\end{thebibliography}

\end{document}